\documentclass{article}

\PassOptionsToPackage{numbers,compress}{natbib}
\usepackage[preprint]{nips_template/neurips_2026}

\usepackage[utf8]{inputenc}
\usepackage[T1]{fontenc}
\usepackage{hyperref}
\usepackage{url}
\usepackage{microtype}
\usepackage{graphicx}
\usepackage[percent]{overpic}
\usepackage{subcaption}
\usepackage{booktabs}
\usepackage{multirow}
\usepackage[table]{xcolor}
\usepackage{xspace}
\usepackage{amsmath}
\usepackage{amssymb}
\usepackage{mathtools}
\usepackage{amsthm}
\usepackage{bbm}
\usepackage{tcolorbox}
\usepackage{placeins}
\usepackage{afterpage}
\usepackage[capitalize,noabbrev]{cleveref}

\definecolor{highlight}{HTML}{F5F5F5}
\definecolor{best}{HTML}{DAE8FC}
\definecolor{second}{HTML}{FCE5CD}
\definecolor{blockgrey}{HTML}{F6F6F6}

\colorlet{PurpleBlue}{purple!20!blue!60}

\newcommand{\yang}[1]{}
\newcommand{\cz}[1]{}
\newcommand{\shortname}{LACE\xspace}

\theoremstyle{plain}

\theoremstyle{definition}

\theoremstyle{remark}

\title{LACE: Lattice Attention for Cross-thread Exploration}

\author{%
  Yang Li \\
  Department of Computer Science \\
  Rutgers University \\
  \And
  Zirui Zhang \\
  Department of Computer Science \\
  Rutgers University \\
  \And
  Yang Liu \\
  Amazon AGI Labs \\
  \And
  Chengzhi Mao \\
  Department of Computer Science \\
  Rutgers University
}

\begin{document}

\maketitle

\begin{abstract}
Current large language models reason in isolation. Although it is common to sample multiple reasoning paths in parallel, these trajectories do not interact, and often fail in the same redundant ways. We introduce LACE, a framework that transforms reasoning from a collection of independent trials into a coordinated, parallel process. By repurposing the model architecture to enable cross-thread attention, LACE allows concurrent reasoning paths to share intermediate insights and correct one another during inference. A central challenge is the absence of natural training data that exhibits such collaborative behavior. We address this gap with a synthetic data pipeline that explicitly teaches models to communicate and error-correct across threads. Experiments show that this unified exploration substantially outperforms standard parallel search, improving reasoning accuracy by over 7 points. Our results suggest that large language models can be more effective when parallel reasoning paths are allowed to interact.
\end{abstract}
\afterpage{%
\begin{figure*}[!t]
	\centering
	\begin{minipage}[t]{0.498\textwidth}
		\centering
		\begin{overpic}[width=\linewidth]{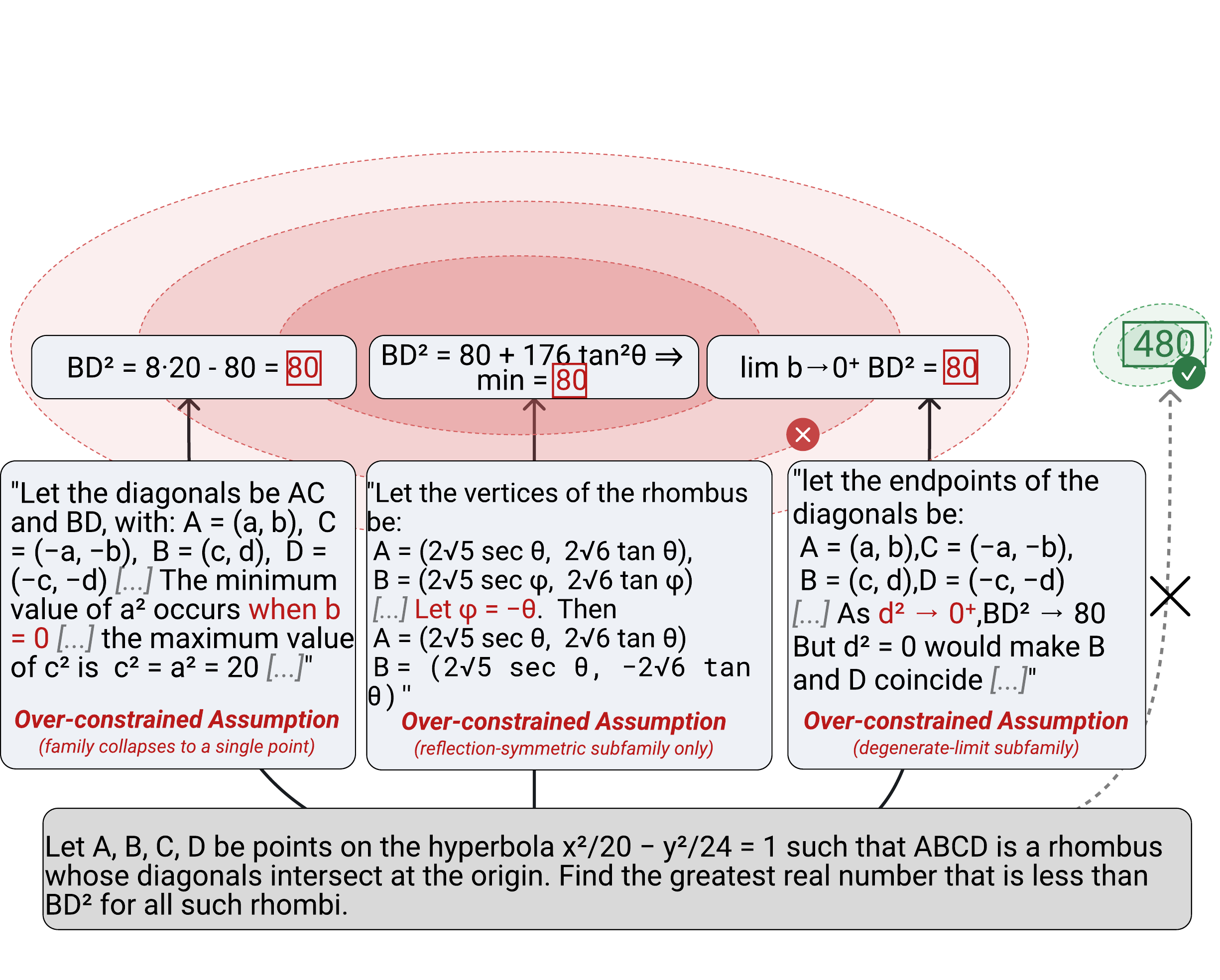}
		\end{overpic}
		\\[-0.2em]
		{\small \textbf{(a)} Isolated samples fall into correlated failure modes.}
	\end{minipage}\hfill
	\begin{minipage}[t]{0.498\textwidth}
		\centering
		\begin{overpic}[width=\linewidth]{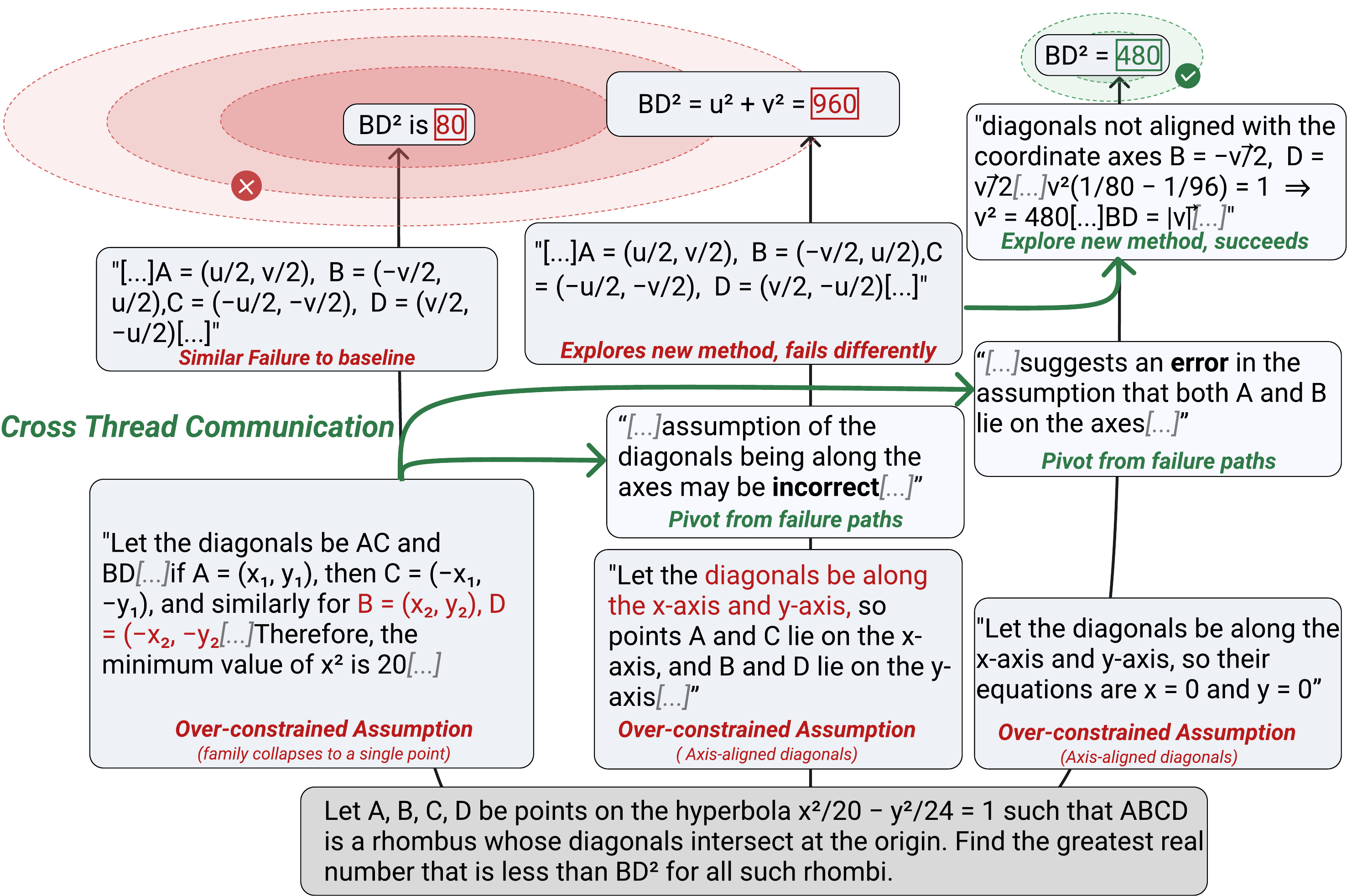}
		\end{overpic}
		\\[-0.2em]
		{\small \textbf{(b)} \shortname explores diverse solutions.}
	\end{minipage}
	\caption{\textbf{Overcoming correlated failures in  reasoning spaces.} 
    For complex problems, correct reasoning paths are often rare and widely separated. 
    \textbf{(a) Isolated Sampling:} Standard parallel generation methods independently explore the space, frequently collapsing into \textbf{correlated failure modes} driven by shared, over-constrained assumptions. 
    \textbf{(b) \shortname (Ours):} By introducing \textbf{latent cross-thread communication} during generation, our method enables concurrent threads to actively pivot away from known failure paths. This shared context encourages the exploration of diverse reasoning modes, significantly increasing the probability of discovering the correct solution.
    }
	\label{fig:teaser}
	\vspace{-0.8em}
\end{figure*}
}
\section{Introduction}

Large Language Models achieve strong performance over a number of reasoning tasks, yet they remain constrained by strictly sequential, left-to-right decoding \cite{vaswani2017attention}. When tackling difficult proofs or complex plans, standard generation models commit to a single path in a vacuum. These limitations prevent the application of language models to highly complex problem-solving scenarios, underscoring the gap between current decoding algorithms and human-level abilities, which rely on collateral thinking to explore multiple hypotheses at once \cite{kahneman2011thinking, johnson2010mental, kay2020constant}.

A large body of work has studied how to scale test-time compute such that models can solve complex reasoning tasks. Parallel sampling and its variants \cite{cobbe2021training, zheng2025parallelR1, wu2025native}, including generating $k$ independent solutions and selecting the best one post-hoc, significantly improve reasoning performance. However, while existing methods focus on increasing the number of samples, they are burdened because they isolate each thread like $k$ agents solving a puzzle in separate rooms. Independent sampling cannot adapt to the breakthroughs or dead-ends of other threads during generation, causing computation to frequently collapse into redundant, correlated errors \cite{kim2025correlated, madaan2023self}.

In this paper, we introduce an approach for \emph{bridging} the generation process, allowing us to formulate a decoding strategy that adapts and collaborates across threads. Our approach, \shortname (Lattice Attention for Cross-thread Exploration), enables threads to share in-situ insights to solve problems collaboratively. \cref{fig:teaser} shows how our method explores diverse solutions compared to isolated sampling. However, multi-threaded reasoning is more challenging to produce than standard sampling because parallel, logically diverse traces are scarce in standard pre-training corpora.

Our key insight is that multi-thread communication must occur directly within the latent space to be efficient and effective. Rather than relying on isolated sequence processing or slow text-based coordination, we design and inject explicit parallel attention pathways, expanding the architecture into a 2-D Lattice Attention structure. This introduces a width dimension that actively shares continuous representations across trajectories. We found that by curating synthetic data with explicit interaction points, we can capitalize on our lattice attention layers to teach models how to collaborate. By using Supervised Fine-Tuning (SFT) and Reinforcement Learning (RL) to train this structure, the model explicitly diversifies search strategies and prevents redundant failures, providing a much more efficient search than post-hoc verification.

A key advantage of our framework is that it transforms reasoning from a set of isolated, independent events into a unified, collaborative exploration. Moreover, because this coordination occurs within the latent space, our approach can perform in-situ evaluation to identify and select the optimal solution on the fly, entirely eliminating the bottleneck of decoding to natural language. Furthermore, this multi-threaded formulation allows us to seamlessly apply diversity rewards directly during RL, incentivizing the threads to explore distinct reasoning paths and maximize overall accuracy.

Visualizations and empirical experiments show that our collateral reasoning strategy significantly improves reasoning diversity and problem-solving capabilities for several established benchmarks. Our method advances  independent sampling baselines by a large margin across challenging reasoning datasets including AIME 25 \cite{aime25}, AIME 24 \cite{aime24}, and LiveBench \cite{livebench} (by up to 13 points). In addition, our empirical results demonstrate that our method learns to share insights across threads, allowing it to pick the best solution on the fly and dramatically reduce wasted compute.

\section{Related Work}

\paragraph{External Test-Time Search and Scaling.} The paradigm of large language models is shifting from scaling parameters to scaling inference-time compute~\citep{snell2024scaling, wu2024inference}, with recent studies showing that additional reasoning-time computation, often framed as \textit{System 2 thinking}, can yield gains comparable to scaling model size~\citep{li2025system}. Existing methods largely realize this extra compute through external search or orchestration, including sequential reasoning strategies such as Chain-of-Thought and its variants~\citep{wei2022chain, kojima2022large, zhou2022least}, structured search methods such as Tree-of-Thoughts~\citep{yao2023tree}, and parallel sampling pipelines based on self-consistency~\citep{wang2022self}, bootstrapped Best-of-$N$ aggregation~\citep{rakhsha2025majority}, learned verifiers or reward models~\citep{stiennon2020learning, cobbe2021training}, RL-based parallel reasoning frameworks such as Parallel-R1 and Native Parallel Reasoner~\citep{zheng2025parallelR1, wu2025native}, and width-oriented external parallel thinking such as ParaThinker~\citep{wen2025parathinker}. Despite their differences, these approaches coordinate reasoning primarily outside the model's native token-level generation process, and remain vulnerable to biased or correlated failures across independently sampled trajectories~\citep{ichihara2025evaluation, huang2023large, madaan2023self}.
\paragraph{Generation-Time Parallel Reasoning.} More recent work has begun to move parallel reasoning into the generation process itself. ParaDecodeOneSeq~\citep{yu2025accelerate} packs multiple branches into a single sequence for efficiency-oriented decoding. Hogwild!~\citep{rodionov2025hogwild} enables \textit{explicit} concurrent attention through shared cross-thread KV states, while GroupThink~\citep{hsu2025group} studies token-level collaboration among concurrent reasoning agents via direct cross-thread interaction. While \shortname shares the goal of generation-time cross-thread interaction with these methods, it differs in mechanism. Rather than exposing each thread to token histories of other threads through shared KV states and non-standard masks, it introduces \emph{implicit} cross-thread interaction through a lightweight gated side path over standard attention-derived representations. This preserves the standard causal-attention backbone while allowing threads to influence one another during generation, turning independent sampling into a collaborative process that explicitly targets redundant exploration and correlated errors~\citep{kim2025correlated, hong2004groups}.

\section{Method}
The central idea of \shortname is to make parallel reasoning more exploratory by allowing threads to collaborate while they are generated. For difficult reasoning problems, correct solutions can occupy sparse modes in the solution space. Independent parallel sampling often revisits nearby failure modes, so simply increasing the number of samples may not increase useful coverage. We therefore build \shortname around three connected requirements. First, the model needs training data in which different threads explore distinct but comparable solution paths. Second, the model needs an architectural channel that lets these threads exchange information during generation. Third, post-training should strengthen the communication and self-selection behaviors that turn diverse candidates into a better final answer. Following this logic, we first seed multi-thread data for complementary exploration in \cref{sec:data}, then connect threads with Lattice Attention in \cref{sec:lace}, and finally post-train the model for collaborative search in \cref{sec:training}.

\subsection{Seeding Multi-Thread Data for Complementary Exploration}
\label{sec:data}

\begin{figure*}[t]
    \centering
    \includegraphics[width=0.98\textwidth]{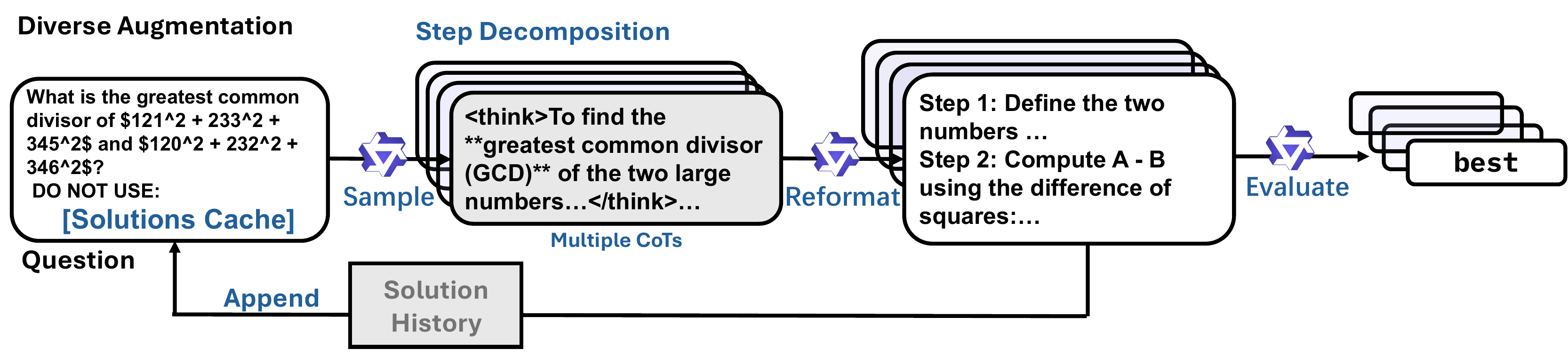}
    \caption{Data curation pipeline for seeding complementary reasoning threads.}
    \label{fig:data_pipeline}
    \vspace{-0.3em}
\end{figure*}

Collaborative reasoning is only meaningful when parallel threads have something useful to share. The training data must therefore satisfy two properties. First, \textbf{reasoning diversity}: each problem should admit multiple logically distinct solution paths, so threads cover different regions of the solution space rather than producing superficial rephrasings. Second, \textbf{cross-thread comparability}: the threads should solve the same problem in a shared format, so the model can compare partial progress, recognize successful paths, and avoid repeating known failures. Driven by these criteria, we synthesize multi-thread training data through the pipeline shown in Figure~\ref{fig:data_pipeline}.

\paragraph{Model-Specific Filtering.}
Not all problems are suitable for multi-thread training. Problems at the extremes of difficulty naturally fail to satisfy our criteria. Overly simple problems typically admit a single obvious solution, which the model tends to sample repeatedly with limited diversity. Conversely, excessively difficult problems prevent the model from sampling any correct solution, thereby lacking positive supervision. Ideal training data for a specific base model requires a balanced mixture of successful and failed reasoning paths~\citep{bengio2009curriculum}, since a failed exploration strategy in one thread can still provide useful information to another. We therefore retain only problems where the base model can produce both correct and incorrect solutions with success rate $S \in (0,0.5]$ across multiple sampling attempts. This upper bound focuses training on problems where failures are at least as common as successes, making cross-thread recovery and comparison useful.

\paragraph{Diverse Thread Augmentation.}
To ensure reasoning diversity across threads, we employ an iterative sampling strategy that explicitly discourages redundant solution paths, inspired by DeRL~\citep{anderl}. For each problem, we maintain a \textit{solution history} that accumulates compressed summaries of previously generated reasoning traces. When sampling a new thread, we prepend this history to the prompt with an explicit instruction to avoid the catalogued approaches, for example \texttt{DO NOT USE: [Solutions Cache]}. This steers the model toward unexplored strategies and implements a soft rejection sampling procedure over the space of solution methods. Each newly sampled trace is then summarized and appended to the solution history, progressively expanding the set of forbidden paths for subsequent threads. We repeat this process until we obtain $T$ diverse reasoning trajectories per problem. To control sequence length for efficient training, we further apply step decomposition that compresses verbose traces (${>}8000$ tokens) into concise step-by-step formats (${<}3000$ tokens) while preserving the core logical flow.

\paragraph{Self-Selection Formatting.}
To make diversity useful for collaboration, we design a self-selection task that requires cross-thread comparison. Given $T$ parallel reasoning paths, the model must identify the thread yielding the best solution. This task encourages the model to use peer information rather than rely on isolated single-thread reasoning. To construct supervision, we use LLM-as-a-judge~\citep{zheng2023judging} to evaluate each thread's validity and logical soundness within the multi-thread context, providing a critique and assigning a grade. The final synthetic training data is formatted as:
\vspace{-0.1em}
\begin{tcolorbox}[colback=gray!5, colframe=gray!50, boxrule=0.3pt, boxsep=0.5mm, left=1mm, right=1mm, top=0.6mm, bottom=0.6mm, width=\columnwidth]
\scriptsize
[question]\\
\hspace*{0.6em}$_t$: [exploration] [comment] [\textsc{tag}],\ $t=1,\ldots,T$
\end{tcolorbox}
\vspace{-0.2em}
where $t$ denotes the thread index, \emph{exploration} contains the reasoning trace, \emph{comment} holds the judge's comparative evaluation, and \emph{tag} indicates the final score. The tag is drawn from \textsc{best}, \textsc{success}, and \textsc{fail}, with exactly one thread designated as \textsc{best}. Both the comment and tag are only well-defined through cross-thread comparison, directly training \shortname to attend to peer threads without an extra LLM-based selection stage.

\subsection{Connecting Threads with Lattice Attention}
\label{sec:lace}

\begin{figure*}[t]
  \centering
  \begin{subfigure}[t]{0.48\linewidth}
    \centering
    \includegraphics[width=\linewidth]{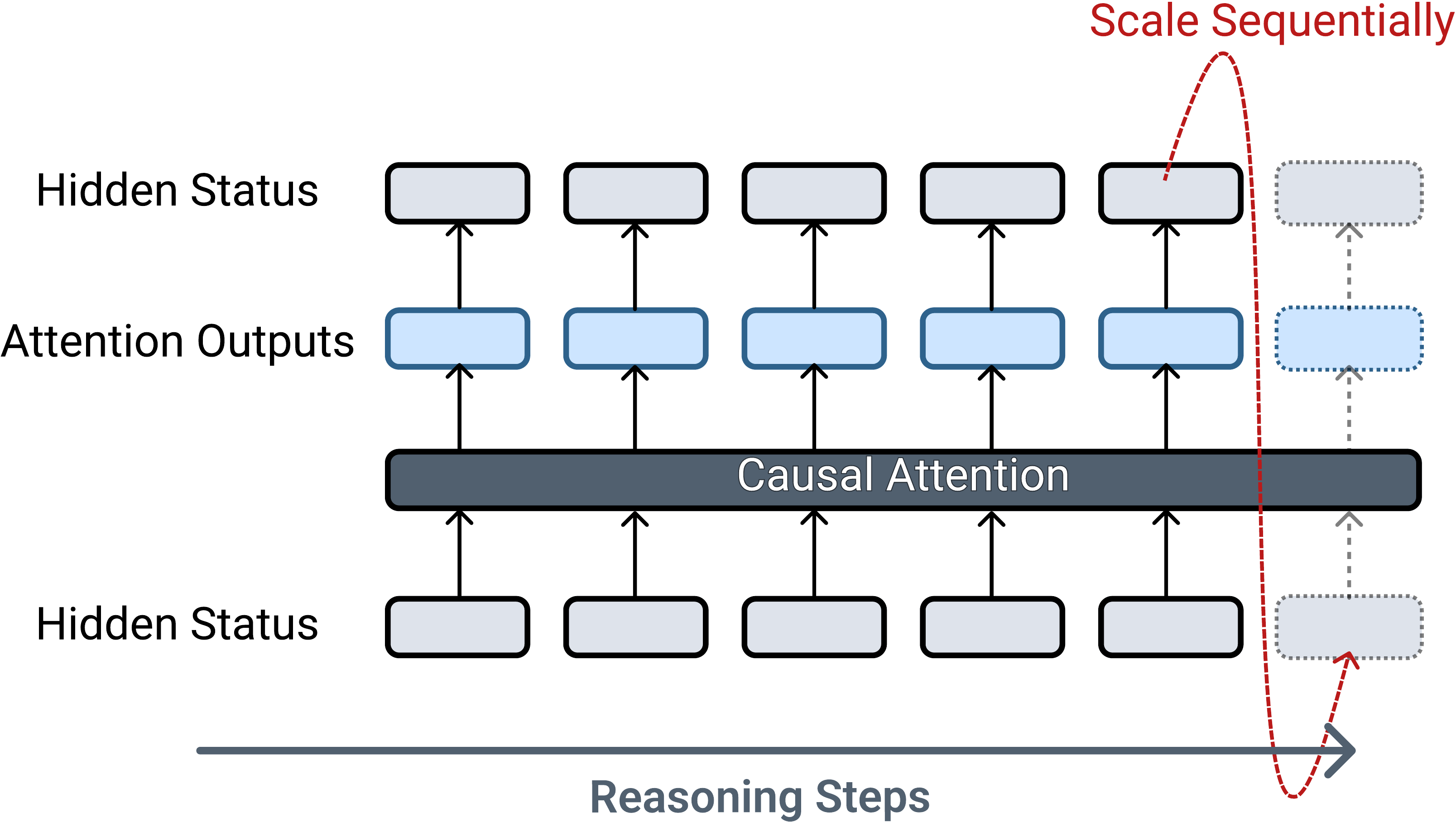}
    \caption{Baseline LLM}
  \end{subfigure}\hfill
  \begin{subfigure}[t]{0.48\linewidth}
    \centering
    \includegraphics[width=\linewidth]{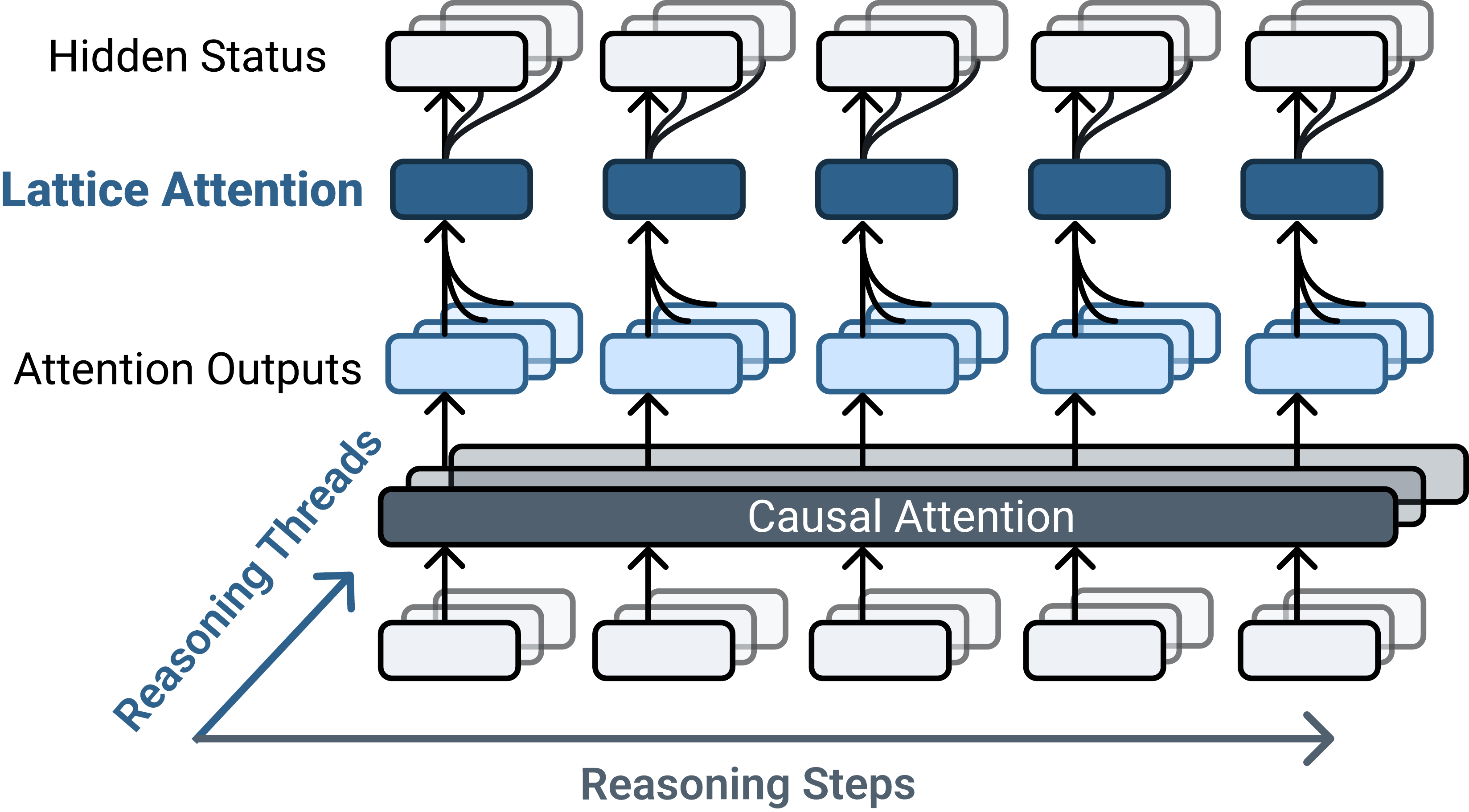}
    \caption{LACE (ours)}
  \end{subfigure}
  \caption{Comparison between the standard parallel decoding pipeline and LACE (ours). Left: a baseline LLM generates reasoning threads independently, so intermediate progress cannot be shared across threads during decoding. Right: LACE (ours) augments the backbone with lattice attention, enabling cross-thread communication through a gated residual path while preserving causal generation within each thread.}
  \label{fig:pipeline}
        \vspace{-0.4em}
\end{figure*}

Given diverse and comparable threads, the next challenge is to let them communicate during generation. We introduce Lattice Attention as a lightweight side path that extends standard causal processing with cross-thread information flow. The mechanism is designed to preserve the pre-trained causal backbone while adding a controlled channel for peer context. Let $B$ denote the batch size, $N$ the number of threads, $L$ the context length, and $D$ the hidden dimension. We denote the head dimension as $d$ and use subscript $p$ for lattice attention components. An overview is shown in Figure~\ref{fig:pipeline}.

\paragraph{Context-Aware Design.}
A naive design would apply cross-thread attention directly over raw token embeddings or early hidden states. However, reasoning is inherently contextual: whether a token is useful to another thread depends on the partial derivation that precedes it, not just on the token itself. Tokens at aligned positions across threads may therefore be poorly matched before each thread has formed enough local reasoning context. This makes naive cross-thread attention both noisy and hard to train. It also increases computation if attention is extended over all thread-token pairs, and large extra layers trained from scratch may disturb well-learned causal attention layers, especially with limited collaborative data. We therefore use a \textbf{context-aware} and \textbf{parameter-efficient} design. Instead of attending over raw embeddings, we operate on the output of standard scaled dot-product attention (SDPA), $\mathbf{A}_{\text{std}} \in \mathbb{R}^{(BN) \times L \times D}$, which already encodes each thread's causal context. This lets \shortname exchange richer intermediate reasoning states while avoiding redundant computation.

To keep the added module lightweight, we use three strategies. First, SDPA outputs are projected to a lower-dimensional lattice space before cross-thread attention. Second, inspired by ControlNet~\citep{zhang2023controlnet}, we selectively plug lattice attention layers into middle-to-last layers of the base model, where cross-thread communication is most useful for complex reasoning~\citep{yang2024large}. Third, a learned gate modulates the contribution of the lattice path, allowing the model to balance thread-independent and thread-aware processing. Together, these choices limit additional parameters to less than $1$\% of the original model while enabling cross-thread information exchange with minimal disruption to the pre-trained causal layers.
\vspace{-0.3em}

\paragraph{Cross-Thread Attention.}
We project the standard attention output into a low-dimensional lattice space, $\mathbf{Z} = \mathbf{A}_{\text{std}} \mathbf{W}_{\text{down},p}$, and compute parallel queries, keys, and values, $\mathbf{Q}_p, \mathbf{K}_p, \mathbf{V}_p = \text{Proj}(\mathbf{Z})$. To encode both token and thread indices, we apply 3D RoPE~\citep{su2024roformer, ma20253d}, yielding $\tilde{\mathbf{Q}}_p, \tilde{\mathbf{K}}_p = \text{RoPE}_{3D}(\mathbf{Q}_p, \mathbf{K}_p; l, n)$, where the first $d_t$ head dimensions encode token position $l$ and the remaining $d_b = d - d_t$ dimensions encode thread index $n$. Cross-thread attention is applied only among SDPA representations at the same decoding step, so a token at position $l$ can exchange information with peer threads at position $l$ after each thread has already summarized its own causal prefix through SDPA. It does not attend to raw future tokens from other threads. We then group the $N$ same-step representations within each batch item and perform cross-thread attention, $\mathbf{A}_p = \text{Attention}(\tilde{\mathbf{Q}}_p, \tilde{\mathbf{K}}_p, \mathbf{V}_p)$. Following Gated Attention~\citep{qiu2025gated, dauphin2017language, shazeer2020glu}, we fuse the standard and lattice paths with a learned gate, $\mathbf{G} = \sigma\left(\text{Linear}([\mathbf{H}_{\text{down}}; \mathbf{A}_p])\right)$ and $\mathbf{O} = \mathbf{A}_{\text{std}} \mathbf{W}_O + (\mathbf{G} \odot \mathbf{A}_p) \mathbf{W}_{O,p}$, where $\mathbf{H}_{\text{down}} = \mathbf{H} \mathbf{W}_{\text{down},h}$ denotes the projected input hidden states.

\subsection{Post-Training LACE for Collaborative Search}
\label{sec:training}

The data pipeline defines what collaborative behavior should look like, and Lattice Attention provides the channel through which it can happen. The final step is to train the model so this channel becomes useful rather than passive. Our training framework consists of supervised finetuning and reinforcement learning.

\paragraph{Supervised Finetuning.}
We initialize {\shortname} from a pre-trained base model and finetune it on our curated multi-thread reasoning data~\citep{ouyang2022training}. To stabilize the newly inserted lattice attention pathway, the initial SFT steps update only the lattice attention parameters using the standard cross-entropy loss~\citep{radford2019language}, adapting cross-thread communication while preserving the base model's causal reasoning ability. We then continue with full-parameter finetuning on the same data to teach the model the multi-thread reasoning format introduced in \cref{sec:data}, including summaries and self-selection tags. We additionally apply random thread shuffling during training with a fixed probability threshold. This prevents the model from overfitting to fixed thread identities and encourages it to rely on cross-thread information rather than reasoning solely from intra-thread context.
\vspace{-0.2em}

\paragraph{Lattice GRPO.}
Finally, we optimize the model with Lattice GRPO, our multi-thread extension of GRPO~\citep{guo2025deepseek, schulman2017proximal, shao2024deepseekmath}. Following baseline GRPO, the policy model rolls out $G$ reasoning groups. In Lattice GRPO, each group contains $T$ thread completions generated simultaneously, enabling reward computation over the full thread group. Along each thread, we prompt the model to self-assess its solution enclosed with \texttt{<summary>...</summary>} and tag each thread with one of \textsc{best}, \textsc{success}, and \textsc{fail}. This turns RL into a direct pressure on both exploration and communication, since the model must generate diverse candidates and identify which thread is most reliable.
\vspace{-0.2em}

\paragraph{Thread-Aggregated Reward.}
For a prompt group containing $T$ parallel threads, $\mathbf{x} = (x^{(1)}, \ldots, x^{(T)})$, the policy samples $G$ reasoning groups. Each group $\mathbf{y}^{(g)} = (y^{(g,1)}, \ldots, y^{(g,T)})$ consists of $T$ thread outputs produced in parallel through lattice attention. Unlike standard GRPO that computes rewards independently for each output, Lattice GRPO first aggregates rewards across the threads in each group and then computes the group-level advantage. The group reward is $r^{(g)} = R_{\text{acc}}(\mathbf{x}, \mathbf{y}^{(g)}) + \lambda_{\text{div}} R_{\text{div}}(\mathbf{y}^{(g)})$, where $R_{\text{acc}}$ is the accuracy reward and $R_{\text{div}}$ is the diversity reward. The accuracy reward encourages the model to evaluate candidate threads and select a reliable best path through self-selection tags. Let $\mathcal{B} = \{t : \text{\textsc{best}} \in y^{(t)}\}$ be the set of threads marked as best, and let $\mathcal{S} = \mathcal{B} \cup \{t : \text{\textsc{success}} \in y^{(t)}\}$ include threads marked as successful. We define $R_{\text{acc}}(\mathbf{x}, \mathbf{y}) = \frac{1}{|\mathcal{S}|}\sum_{t \in \mathcal{S}} \mathbbm{1}[\text{verify}(y^{(t)}, a^{(t)})]$ when the self-selection tags are present, and assign a negative reward to malformed outputs. Here, $a^{(t)}$ is the ground-truth answer for thread $t$, and $\text{verify}(\cdot, \cdot)$ performs symbolic mathematical verification. This reward promotes self-selection and cross-thread evaluation without requiring an external judge at inference time. To encourage diverse reasoning trajectories across threads, we additionally define a diversity reward based on embedding dissimilarity~\citep{zhang2025qwen3}, $R_{\text{div}}(\mathbf{y}) = \frac{2}{T(T-1)} \sum_{i < j} \left(1 - \cos(\mathbf{e}^{(i)}, \mathbf{e}^{(j)})\right)$, where $\mathbf{e}^{(t)} = \text{Embed}(y^{(t)})$ is computed by prompting a pre-trained embedding model to represent the logical path extracted from the reasoning steps. This reward reaches its maximum when all threads explore distinct reasoning paths.

Crucially, the advantage $A^{(g)}$ is broadcast to all $T$ threads within generation $g$, i.e., $A^{(g,t)} = A^{(g)}$ for all $t \in \{1, \ldots, T\}$. This shared advantage signal encourages all threads to collectively improve, reinforcing the collaborative nature of multi-thread reasoning.

\paragraph{Lattice GRPO Objective.}
The final Lattice GRPO objective combines the clipped surrogate loss with a KL penalty:
\begin{equation}
    \resizebox{0.98\linewidth}{!}{$\displaystyle \mathcal{L}_{\text{Lattice-GRPO}}(\theta) = -\mathbb{E}\!\left[\sum_{g=1}^G \sum_{t=1}^T \frac{1}{L^{(g,t)}} \sum_{l=1}^{L^{(g,t)}} \min\!\left(\rho_l^{(g,t)} A^{(g)}, \text{clip}(\rho_l^{(g,t)}, 1-\epsilon, 1+\epsilon) A^{(g)}\right)\right] + \beta D_{\text{KL}}$}
\end{equation}
where $\rho_l^{(g,t)}$ is the importance sampling ratio for token $l$ in thread $t$ of group $g$, with the policy distribution conditioned on the thread's own prefix and peer same-step lattice states. $L^{(g,t)}$ is the completion length, and $D_{\text{KL}}$ is the KL divergence from a reference policy.

\begin{figure*}[t]
  \centering
  \makebox[\textwidth][c]{%
  \begin{subfigure}[b]{0.32\textwidth}
    \centering
    \includegraphics[width=\linewidth]{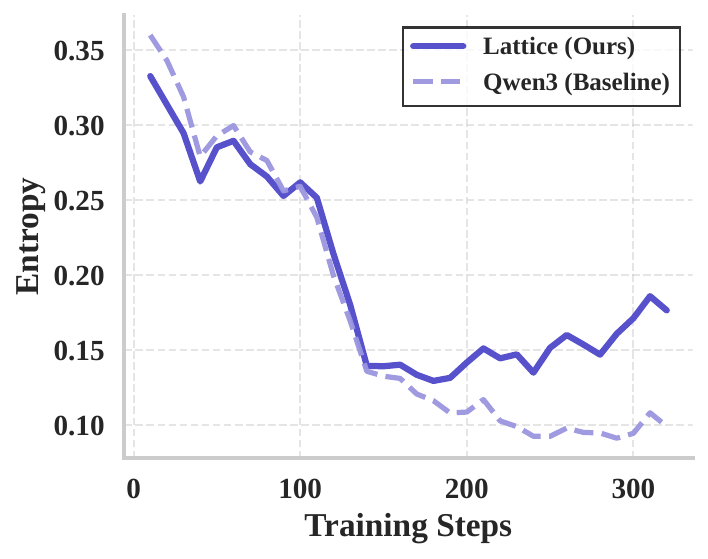}
    \caption{Entropy}
    \label{fig:entropy}
  \end{subfigure}\hfill
  \begin{subfigure}[b]{0.32\textwidth}
    \centering
    \includegraphics[width=\linewidth]{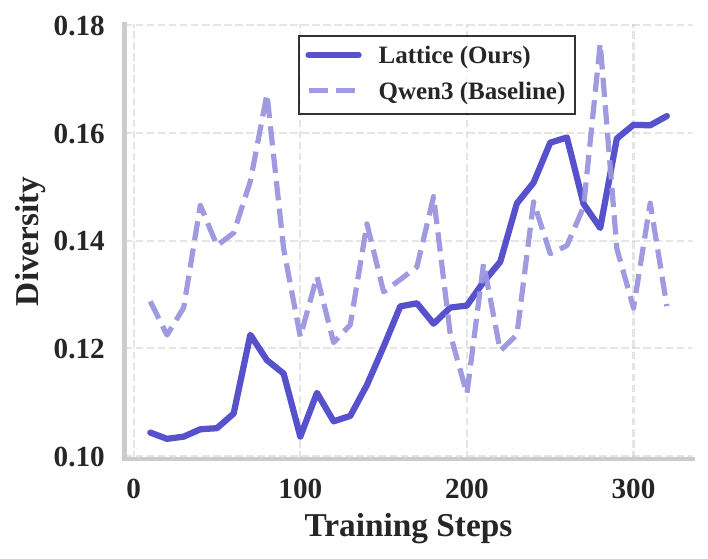}
    \caption{Diversity}
    \label{fig:diversity}
  \end{subfigure}\hfill
  \begin{subfigure}[b]{0.32\textwidth}
    \centering
    \includegraphics[width=\linewidth]{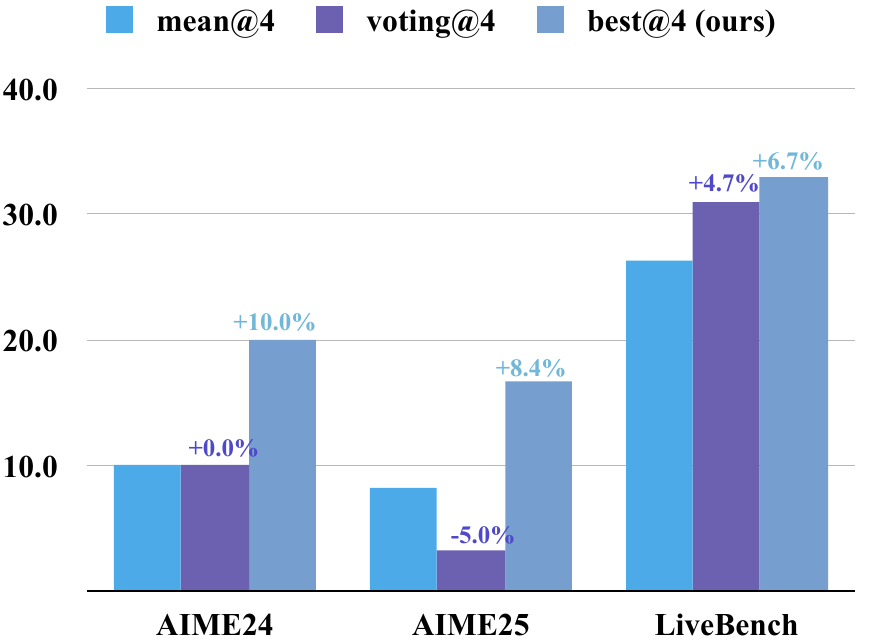}
    \caption{Self-selection vs. voting.}
    \label{fig:mean_vs_best}
  \end{subfigure}%
  }
  \captionsetup{width=\textwidth}
  \caption{\textbf{LACE improves self-selection by fostering more effective exploration.} Panels (\textbf{a}) and (\textbf{b}) compare the entropy and diversity dynamics of Lattice-4B against the Qwen3-4B baseline during RL training: while the baseline suffers from entropy collapse after roughly 150 steps, Lattice-4B exhibits a stable entropy rebound and continuously increasing diversity. Panel (\textbf{c}) compares Mean, Voting, and self-selected Best predictions under a strict self-selection diagnostic, where Best is counted only when the model emits exactly one \textsc{best} thread. This verifies whether the model can select a single reliable candidate rather than merely generate multiple answers.}
  \label{fig:exploration_comparison}
  \vspace{-0.4em}
\end{figure*}

\section{Experiments}
We conduct experiments to evaluate the effectiveness of our proposed \shortname framework on challenging math and reasoning benchmarks. We compare our method against strong baselines, assessed by accuracy and exploration diversity in \cref{sec:main-results}. We then analyze its \textbf{emergent cross-thread mechanisms}~\citep{wei2022emergent, woolley2010evidence} and ablate the data and training pipeline in \cref{sec:qualitative_analysis}, showing how collaborative exploration and self-regulated behaviors arise from the lattice architecture.
\vspace{-0.4em}
\paragraph{Implementation Details.} We implement our method based on base models Qwen3~\citep{yang2025qwen3} with 1.7B and 4B parameters. Our training data is curated from the DAPO dataset~\citep{yu2025dapo} using our data curation pipeline described in \cref{sec:data}. Four threads are applied in the lattice structure for all experiments. Lattice attention layers are inserted every other layer, starting from the middle to the last original layer. Detailed thread-scaling micro-benchmarks are reported in Appendix~\cref{appendix:overhead}, showing negligible FLOPs overhead ($<1.3\%$), modest memory overhead (roughly $1.6$--$15\%$), and latency overhead that is primarily memory-bandwidth bound. The detailed training hyperparameters and data statistics are provided in Appendix~\cref{appendix:data_stats,appendix:implementation}.
\vspace{-0.2em}
\subsection{Quantitative Results}
\label{sec:main-results}
\paragraph{Baselines and Metrics.} We compare our \shortname framework against two strong baselines: \textbf{Independent} and \textbf{Isolated Parallel}. The Independent baseline is a standard single-thread model trained with conventional SFT and RL; at inference time, we sample multiple independent trajectories and report both mean accuracy and majority voting. The Isolated Parallel baseline uses the same multi-thread data format and SFT/RL recipe as \shortname, but keeps the standard base model without inserted lattice layers or cross-thread attention, so the threads are trained and decoded in parallel yet remain independent. To ensure a fair comparison, all methods undergo SFT and RL stages; for \shortname, the initial SFT steps are used to initialize the newly inserted lattice layers before standard full-parameter finetuning. Appendix~\cref{appendix:baseline_ext} additionally compares against judge-based selection~\citep{zheng2023judging} and sequential refinement~\citep{madaan2023self} baselines, showing that they become competitive only with stronger external judges or extra decoding cost, while LACE preserves a better efficiency profile. We evaluate the models using \textbf{Accuracy} and \textbf{Diversity} (Div.).
As described in \cref{sec:data}, \shortname and Isolated Parallel are trained to self-select the optimal thread. For these multi-thread methods, accuracy is computed from the answer in the thread that the model marks as \textsc{best} among four simultaneous threads. For the Independent baseline, which lacks a self-selection mechanism, we report the mean accuracy across samples~\citep{chen2021evaluating}. Additionally, we report majority voting to compare our learned self-selection against a standard post-hoc Best-of-N strategy. \textbf{Diversity} is computed using the embedding dissimilarity reward defined in \cref{sec:training}.

\begin{table*}[t]
\centering
\vspace{-0.5em}
\scriptsize
\caption{\textbf{Comparisons with baselines on math and reasoning benchmarks.} \textbf{Independent} is a single-thread SFT+RL baseline, while \textbf{Isolated Parallel} uses the same multi-thread training recipe as \shortname without lattice layers or cross-thread attention. We report \textbf{Accuracy} and \textbf{Diversity} (Div.). By default, Independent reports Mean@4, while Isolated Parallel and \shortname use the answer from the thread selected by the model as \textsc{best} among four simultaneous threads; +Vote denotes replacing the default selection rule with majority voting.}
\label{tab:main_results}
\setlength{\aboverulesep}{0.25ex}
\setlength{\belowrulesep}{0.35ex}
\setlength{\tabcolsep}{3.6pt}
\renewcommand{\arraystretch}{1.04}
\makebox[\textwidth][c]{%
\resizebox{0.98\textwidth}{!}{%
\begin{tabular}{c@{\hspace{0.45em}}ll cc cc cc}
\toprule
\textbf{Model} & \textbf{Method} & \textbf{Train}
& \multicolumn{2}{c}{\textbf{AIME 25}}
& \multicolumn{2}{c}{\textbf{AIME 24}}
& \multicolumn{2}{c}{\textbf{LiveBench}} \\
\cmidrule[\lightrulewidth](lr){4-5} \cmidrule[\lightrulewidth](lr){6-7} \cmidrule[\lightrulewidth](lr){8-9}
& & & \textbf{Accuracy} & \textbf{Div. $\uparrow$} & \textbf{Accuracy} & \textbf{Div. $\uparrow$} & \textbf{Accuracy} & \textbf{Div. $\uparrow$} \\
\specialrule{0.045em}{0.35ex}{0.6ex}

\multirow[c]{7}{1.7cm}{\centering\raisebox{-1.0ex}{\begin{tabular}[c]{@{}c@{}}\itshape\bfseries Qwen3\\[-0.08em]\itshape\bfseries 1.7B\end{tabular}}}
& \textit{Independent} & SFT $\rightarrow$ RL
& 8.3 & 0.07 & 8.3 & 0.07 & 14.6 & 0.06 \\
& \multicolumn{1}{c}{\textit{+ Vote}} & SFT $\rightarrow$ RL
& 10.0 & 0.07 & 10.0 & 0.07 & 16.0 & 0.06 \\
\cmidrule[\cmidrulewidth](l{0.6em}r{0.6em}){3-9}
& \textit{Isolated Parallel} & SFT
& 0.0 & 0.10 & 0.0 & 0.11 & 9.0 & \textbf{0.11} \\
& & SFT $\rightarrow$ RL
& 6.7 & 0.16 & 3.3 & 0.19 & 20.7 & 0.10 \\
& \multicolumn{1}{c}{\textit{+ Vote}} & SFT $\rightarrow$ RL
& 0.0 & 0.16 & 0.0 & 0.19 & 15.5 & 0.10 \\
\cmidrule[\cmidrulewidth](l{0.6em}r{0.6em}){3-9}

\rowcolor{blockgrey}
\cellcolor{white} & \textit{\textbf{LACE}} & SFT
& 10.0 & 0.10 & 3.3 & 0.11 & 13.7 & 0.11 \\
\rowcolor{blockgrey}
\cellcolor{white} & & SFT $\rightarrow$ RL
& \textbf{13.3}
& \textbf{0.28}
& \textbf{16.7}
& \textbf{0.24}
& \textbf{21.7}
& 0.11 \\

\specialrule{0.06em}{0.45ex}{0.5ex}

\multirow[c]{7}{1.7cm}{\centering\raisebox{-1.0ex}{\begin{tabular}[c]{@{}c@{}}\itshape\bfseries Qwen3\\[-0.08em]\itshape\bfseries 4B\end{tabular}}}
& \textit{Independent} & SFT $\rightarrow$ RL
& 13.3 & 0.06 & 12.5 & 0.07 & 25.6 & 0.15 \\
& \multicolumn{1}{c}{\textit{+ Vote}} & SFT $\rightarrow$ RL
& 13.3 & 0.06 & 13.3 & 0.07 & 28.0 & 0.15 \\
\cmidrule[\cmidrulewidth](l{0.6em}r{0.6em}){3-9}
& \textit{Isolated Parallel} & SFT
& 10.0 & 0.09 & 10.0 & 0.10 & 11.2 & 0.11 \\
& & SFT $\rightarrow$ RL
& 6.7 & 0.10 & 10.0 & 0.10 & 12.7 & 0.15 \\
& \multicolumn{1}{c}{\textit{+ Vote}} & SFT $\rightarrow$ RL
& 3.3 & 0.10 & 10.0 & 0.10 & 15.5 & 0.15 \\
\cmidrule[\cmidrulewidth](l{0.6em}r{0.6em}){3-9}

\rowcolor{blockgrey}
\cellcolor{white} & \textit{\textbf{LACE}} & SFT
& 10.0 & 0.09 & 6.7 & 0.09 & 18.5 & 0.10 \\
\rowcolor{blockgrey}
\cellcolor{white} & & SFT $\rightarrow$ RL
& \textbf{20.0}
& \textbf{0.16}
& \textbf{20.0}
& \textbf{0.16}
& \textbf{41.8}
& \textbf{0.16}
\\

\bottomrule
\end{tabular}
}
}
\vspace{-1.2em}
\end{table*}

\paragraph{Main Results.} As summarized in \cref{tab:main_results}, \shortname consistently achieves superior accuracy across all benchmarks and model scales, with the most significant gains observed after the RL stage. The gains over both Independent Sampling and Isolated Parallel show that the model learns to use cross-thread information for reliable self-selection, rather than merely benefiting from a multi-sample decoding budget. Regarding exploration diversity, training starts from a noisier high-entropy regime because the new cross-thread pathway is randomly initialized, while the subsequent decrease during SFT followed by a sustained rebound during RL indicates genuine, learned exploration rather than stochastic noise. As shown in \cref{fig:exploration_comparison}, this improved exploration expands candidate coverage and, in turn, enables more accurate self-selection than either mean aggregation or majority voting under a strict one-\textsc{best} diagnostic. This combination of high accuracy and meaningful diversity validates the framework's ability to foster collaborative reasoning and enhance the reliability of multi-threaded generation. Beyond math benchmarks, we additionally evaluate Lattice-1.7B on TextWorldCookAgent from TALES~\citep{cui2025tales,cote18textworld}, an interactive agent task with longer-horizon exploration and intermediate feedback. Appendix~\cref{appendix:textworld_generalization} shows that LACE achieves the highest Win Rate among matched agent baselines. Appendix~\cref{appendix:thread_scaling} and \cref{tab:thread_scaling_agent,tab:thread_scaling_livebench,tab:model_scale_preliminary} further show that inference-time scaling remains beneficial without retraining and that preliminary 8B results preserve the same collaborative trend.

\subsection{Analysis and Ablations}
\label{sec:qualitative_analysis}
\label{sec:abla-study}
\vspace{-0.1em}
\paragraph{LACE avoids mode collapse and maintains diverse exploration.} We investigate the training dynamics to understand how the model learns to explore. As shown in \cref{fig:exploration_comparison}, the baseline model suffers from mode collapse, evidenced by a continuous drop in entropy. In contrast, \shortname exhibits a unique pattern where entropy slowly rebounds after an initial decrease, accompanied by a steady rise in diversity scores. This suggests that the synergy between lattice attention and diversity rewards successfully enables \textbf{learned exploration}, allowing the model to actively seek diverse reasoning paths rather than converging to a single mode.
Appendix~\cref{appendix:diversity_sensitivity,tab:diversity_sensitivity} further shows that the diversity reward behaves like a training-time temperature: the default weight provides the best balance between path coverage and focused best-path selection, while larger weights can improve Pass@4 on harder tasks but make stable self-selection harder.
\vspace{-0.4em}
\paragraph{LACE communicates most when threads need peer context.} 
\begin{figure*}[t]
  \centering
  \begin{subfigure}[t]{0.48\linewidth}
    \centering
    \includegraphics[width=\linewidth]{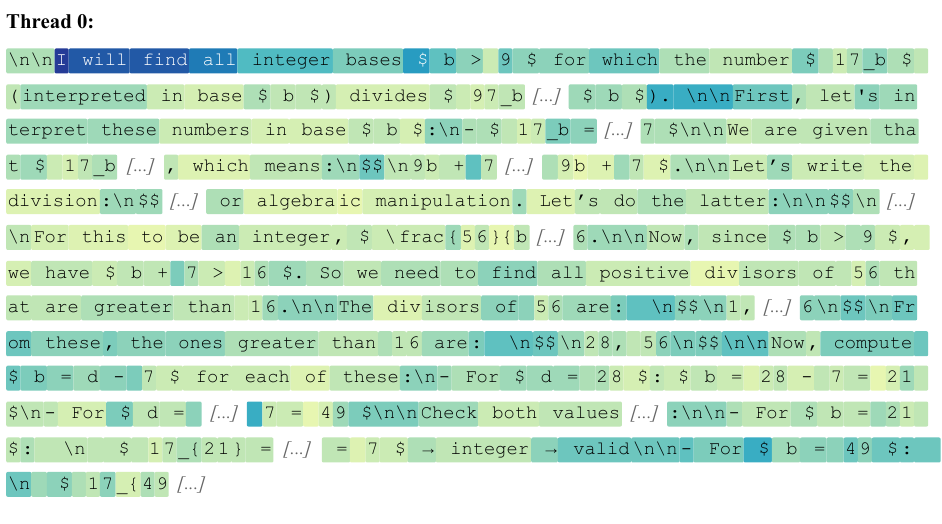}
    \caption{Thread 1}
  \end{subfigure}
  \hfill
  \begin{subfigure}[t]{0.48\linewidth}
    \centering
    \includegraphics[width=\linewidth]{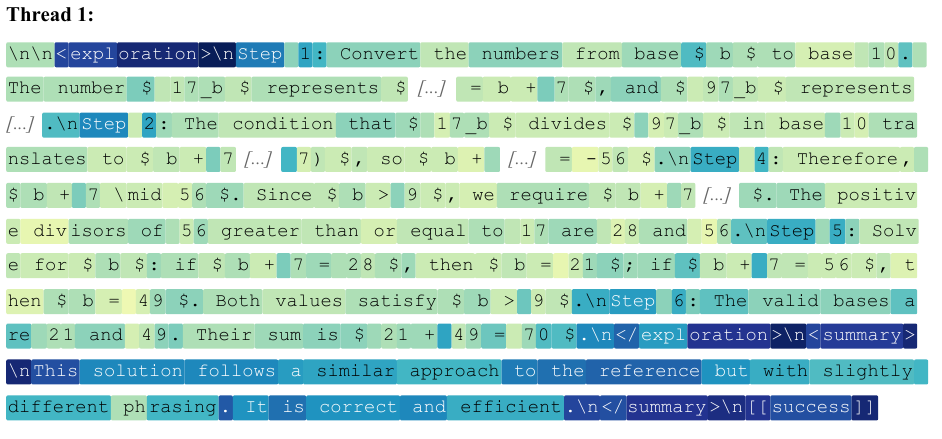}
    \caption{Thread 2}
  \end{subfigure}

  \vspace{0.5em}

  \begin{subfigure}[t]{0.48\linewidth}
    \centering
    \includegraphics[width=\linewidth]{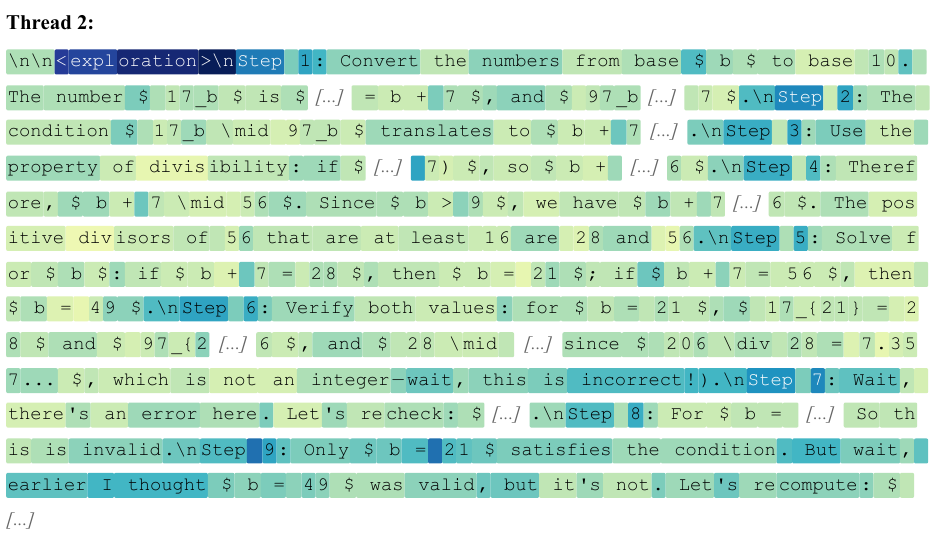}
    \caption{Thread 3}
  \end{subfigure}
  \hfill
  \begin{subfigure}[t]{0.48\linewidth}
    \centering
    \includegraphics[width=\linewidth]{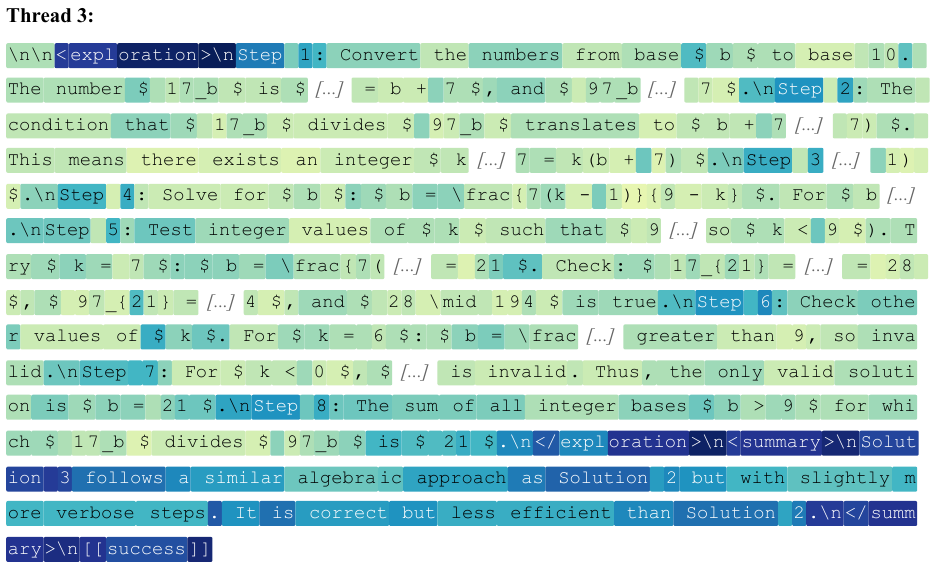}
    \caption{Thread 4}
  \end{subfigure}
  \caption{\textbf{LACE coordinates information sharing across all four threads at critical reasoning steps.} We visualize cross-thread intensity on an AIME 25 example from Lattice-1.7B, where \colorbox{PurpleBlue}{darker shading} indicates stronger cross-thread interaction. Despite no direct supervision, high cross-thread communication in latent space consistently emerges during exploration, self-assessment, and evaluation, and the generated summaries explicitly compare peer solutions.}
  \label{fig:example_generation}
  \vspace{-0.2em}
\end{figure*}
Lattice Attention utilizes a gating mechanism to regulate cross-thread information flow, providing a unique lens to analyze collaborative intensity across both temporal and semantic dimensions. 
Temporally, Appendix~\cref{fig:gate_score} illustrates that information flow peaks during the core reasoning phase, exhibiting a diverge-then-converge strategy coinciding with the classical framework of human intelligence proposed by~\citep{guilford1967nature}, which distinguishes between divergent production and convergent production to arrive at a final correct solution. 
Semantically, \cref{fig:example_generation} reveals that this interaction is content-specific across all four threads: high gate scores align with tokens that require peer information, including exploration markers, summary steps, and self-assessment tags. The generated text provides complementary evidence: for example, Thread 4 explicitly compares its path against other threads in its summary, indicating that the model uses peer trajectories when evaluating candidate solutions.
This suggests that \shortname enables \textbf{need-aware collaboration}: the model increases cross-thread communication when comparing, summarizing, or selecting among candidate paths, rather than exchanging information uniformly throughout generation. Appendix~\cref{sec:gate_stats,appendix:layer_sync} further shows that synchronization concentrates in middle-to-late layers and peaks near Layer 26 across both model scales.
\vspace{-0.3em}
\paragraph{Diversity-aware data and staged training are both necessary.}
\cref{tab:ablation_studies} shows that our curated data produces more diverse threads and stronger cross-thread attention than Parallel-R1 data, while \cref{tab:ablation_studies_expanded} shows that SFT is needed before RL can reliably elicit self-selection.
\begin{table*}[t]
\centering
\scriptsize
\setlength{\aboverulesep}{0.2ex}
\setlength{\belowrulesep}{0.35ex}
\setlength{\tabcolsep}{3.6pt}
\renewcommand{\arraystretch}{1.06}
\makebox[\textwidth][c]{%
\resizebox{0.72\textwidth}{!}{%
\begin{tabular}{l cc c cc c}
\toprule
\textbf{Training Recipe}
& \multicolumn{3}{c}{\textbf{AIME 25}} & \multicolumn{3}{c}{\textbf{LiveBench}} \\
\cmidrule[\lightrulewidth](lr){2-4} \cmidrule[\lightrulewidth](lr){5-7}
& \textbf{Accuracy} & \textbf{Div.} & \textbf{Format}
& \textbf{Accuracy} & \textbf{Div.} & \textbf{Format} \\
\specialrule{0.045em}{0.2ex}{0.35ex}
RL only (w/o SFT) & 0.0 & 0.285 & 26.7 & 1.0 & 0.244 & 31.5 \\
SFT $\rightarrow$ RL w/ Parallel-R1 & 0.0 & 0.203 & 50.0 & 1.5 & 0.148 & 20.5 \\
\rowcolor{blockgrey}
\textbf{SFT $\rightarrow$ RL w/ Our Data} & \textbf{13.3} & \textbf{0.608} & \textbf{73.3} & \textbf{11.5} & \textbf{0.276} & \textbf{36.5} \\
\bottomrule
\end{tabular}%
}
}
\caption{\textbf{Diversity-aware data curation improves strict self-selection accuracy and diversity.} We compare training recipes under the 1.7B ablation protocol with 200 RL steps. Strict accuracy uses the model-selected \textsc{best} thread only when exactly one \textsc{best} marker is produced; Format reports this one-\textsc{best} rate.}
\label{tab:ablation_studies_expanded}
\vspace{-1.0em}
\end{table*}
\begin{table}[t]
\centering
\footnotesize
\setlength{\aboverulesep}{0.35ex}
\setlength{\belowrulesep}{0.55ex}
\setlength{\tabcolsep}{5.0pt}
\renewcommand{\arraystretch}{1.18}
\begin{tabular}{c cc c}
\toprule
\multirow[c]{2}{*}{\raisebox{0.35ex}{\textbf{Data}}} & \multicolumn{2}{c}{\textbf{Diversity $\uparrow$}} & \multirow[c]{2}{*}{\raisebox{0.35ex}{\textbf{\shortstack[c]{Cross-Thread\\Attention $\uparrow$}}}} \\
\cmidrule[\lightrulewidth](lr){2-3}
& \textbf{Edit Distance} & \textbf{Embedding Dissimilarity} & \\
\specialrule{0.045em}{0.35ex}{0.6ex}
Parallel-R1 Data \textit{+ SFT} & 0.13 & 0.00 & 0.307\\
\textbf{Our Data} \textit{+ SFT} & \textbf{0.77} & \textbf{0.12} & \textbf{0.353} \\
\bottomrule
\end{tabular}%
\caption{\textbf{Our curated training data induces both more diverse reasoning paths and stronger cross-thread communication than Parallel-R1 data after SFT.} We compare our curated data with Parallel-R1 data using Edit Distance, Embedding Dissimilarity, and Cross-Thread Attention Ratio, where higher is better for all three metrics.}
\label{tab:ablation_studies}
\vspace{-2.5em}
\end{table}
To measure the intensity of cross-thread communication, we define the cross-thread ratio as
$r_{\text{cross}} = \frac{\bar{A}_{\text{cross}}}{\bar{A}_{\text{self}} + \bar{A}_{\text{cross}}}$,
where $\bar{A}_{\text{self}}$ denotes the average attention weight assigned to keys from the same thread and $\bar{A}_{\text{cross}}$ denotes the average attention weight assigned to keys from peer threads. As shown in \cref{tab:ablation_studies}, our diversity-aware data pipeline yields substantially higher thread diversity and cross-thread attention ratios compared to naïve parallel sampling, validating that data quality is the prerequisite for emergent collaboration. \cref{tab:ablation_studies_expanded} further validates the necessity of the full training recipe: skipping SFT leads to near-zero strict self-selection accuracy and degraded one-\textsc{best} format adherence, indicating that the lattice layers require proper initialization before RL can elicit meaningful cross-thread behaviors. Appendix~\cref{appendix:sa_identity_ablation} and \cref{tab:sa_ablation,tab:path_id_ablation} further show that self-assessment is important for reliable best-thread selection and that combining prompt path IDs with RoPE yields the strongest identity encoding. Appendix~\cref{appendix:thread_scaling} and \cref{tab:model_scale_preliminary} additionally provide preliminary larger-scale results, which suggest that the same collaborative trend persists at 8B.




\vspace{-1.5em}
\section{Conclusion}
\vspace{-0.75em}
Current reasoning methods mainly scale the depth of individual chains, while leaving reasoning paths largely isolated. To address this, we introduce LACE, which adds a lightweight cross-thread side path to transformers and enables \textbf{implicit token-level cross-thread interaction during generation}, rather than post-hoc aggregation or isolated parallel sampling. With synthetic collateral-thinking data and lattice-based RL, LACE turns redundant samples into interactive exploration and delivers consistent gains over matched baselines, including a significant improvement in our core setting. Overall, our findings suggest that stronger reasoning may come not only from stronger individual threads, but also from emergent cross-thread collaboration.

\section{Acknowledgment}
This work is supported by Amazon's Gift Fund.

\bibliographystyle{plainnat}
\bibliography{example_paper}

\newpage
\appendix
\onecolumn

\section{Analytical Results}
\label{sec:analysis}

We characterize the performance gap between independent trace sampling (Best-of-N) and our proposed collateral thinking framework (LACE). We frame the reasoning task as searching for a target solution $y^*$ within a discrete solution space $\mathcal{Y}$. Let $\epsilon_0 = P(y \neq y^* | X)$ be the base error rate of a single model trace.

\textbf{Lemma 1 (Independent Failure Bound).}
In the standard Best-of-N setting, $N$ traces $y_1, \dots, y_N$ are sampled i.i.d. from $P(Y|X)$. The probability that the system fails to uncover $y^*$ is exactly the product of the individual error rates:
\begin{equation}
    P_{IND}(S=0) = \prod_{k=1}^N P(y_k \neq y^*) = \epsilon_0^N
\end{equation}
Because samples are independent, observing that threads $1 \dots k-1$ have failed provides no information to correct thread $k$. The error rate $\epsilon_0$ remains constant for every sample.

\textbf{Theorem 1 (LACE Improvement via Conditional Probability).}
LACE introduces a dependency where each thread can condition on peer latent states at aligned generation steps. Let $E_k$ be the event that thread $k$ fails ($y_k \neq y^*$). The joint failure probability for LACE is given by the chain rule:
\begin{equation}
    \hspace{-1em} P_{LACE}(S=0) = P(E_1) \cdot P(E_2 | E_1) \cdot \dots \cdot P(E_N | E_1, \dots, E_{N-1}) \nonumber
\end{equation}
Assume that cross-thread conditioning helps later threads avoid repeated error modes, so for $k > 1$ the conditional failure probability can be written as $P(E_k | E_{<k}) = \epsilon_0 - \delta_k$, where $\delta_k \geq 0$ captures the error reduction from peer information.

\textit{Proof.}
Let $\mathcal{M}_{err}$ be the set of incorrect high-probability modes (hallucinations or common pitfalls). The base error rate is the sum of probabilities over this set: $\epsilon_0 = \sum_{m \in \mathcal{M}_{err}} P(m)$.
The LACE training objective encourages diverse exploration and self-selection across threads. Under the assumption above, this reduces the probability that thread $k$ repeats an error mode $m$ already represented in peer trajectories. This acts as a soft "rejection sampling" mechanism.

If previous threads have "consumed" a subset of error modes $\mathcal{M}_{visited} \subset \mathcal{M}_{err}$ with total probability mass $\pi_{visited} = \sum_{m \in \mathcal{M}_{visited}} P(m)$, the distribution for thread $k$ is effectively renormalized over the remaining space $\mathcal{Y} \setminus \mathcal{M}_{visited}$.
The new probability of picking the correct answer $y^*$ becomes:
\begin{equation}
    P(y_k = y^* | E_{<k}) \approx \frac{P(y^*)}{1 - \pi_{visited}} > P(y^*)
\end{equation}
Since the probability of success increases, the probability of failure must decrease. Let $\delta_k$ represent this probability mass shift away from the visited error modes. We can rewrite the conditional error rate for the $k$-th thread as:
\begin{equation}
    P(E_k | E_{<k}) = \epsilon_0 - \delta_k
\end{equation}
where $\delta_k$ is a non-negative term scaling with the probability mass of the errors avoided by peers. The total failure probability for LACE becomes:
\begin{equation}
    P_{LACE}(S=0) = \epsilon_0 \cdot (\epsilon_0 - \delta_2) \cdot (\epsilon_0 - \delta_3) \dots (\epsilon_0 - \delta_N)
\end{equation}
Whenever at least one $\delta_k$ is positive, it follows that:
\begin{equation}
    \prod_{k=1}^N (\epsilon_0 - \delta_k) < \epsilon_0^N \implies P_{LACE}(S=1) > P_{IND}(S=1)
\end{equation}
Thus, when cross-thread conditioning reduces repeated error modes, LACE improves the success probability compared to independent sampling under this conditional-error model.

\definecolor{gainlight}{HTML}{EAF4E6}
\definecolor{gainstrong}{HTML}{CFE8CF}
\definecolor{costlight}{HTML}{FBE5E5}
\definecolor{coststrong}{HTML}{F4CCCC}
\definecolor{gaintext}{HTML}{2E5E2E}
\definecolor{costtext}{HTML}{8A1F1F}

\newcommand{\gaincell}[1]{\cellcolor{gainlight}\textcolor{gaintext}{#1}}
\newcommand{\gainmaxcell}[1]{\cellcolor{gainstrong}\textcolor{gaintext}{\textbf{#1}}}
\newcommand{\gainrowcell}[1]{\cellcolor{gainlight}\textcolor{gaintext}{#1}}
\newcommand{\costcell}[1]{\cellcolor{costlight}\textcolor{costtext}{#1}}
\newcommand{\costmaxcell}[1]{\cellcolor{coststrong}\textcolor{costtext}{\textbf{#1}}}

\section{Gate Score Statistics}
\label{sec:gate_stats}
\begin{figure}
    \centering
    \includegraphics[width=0.7\linewidth]{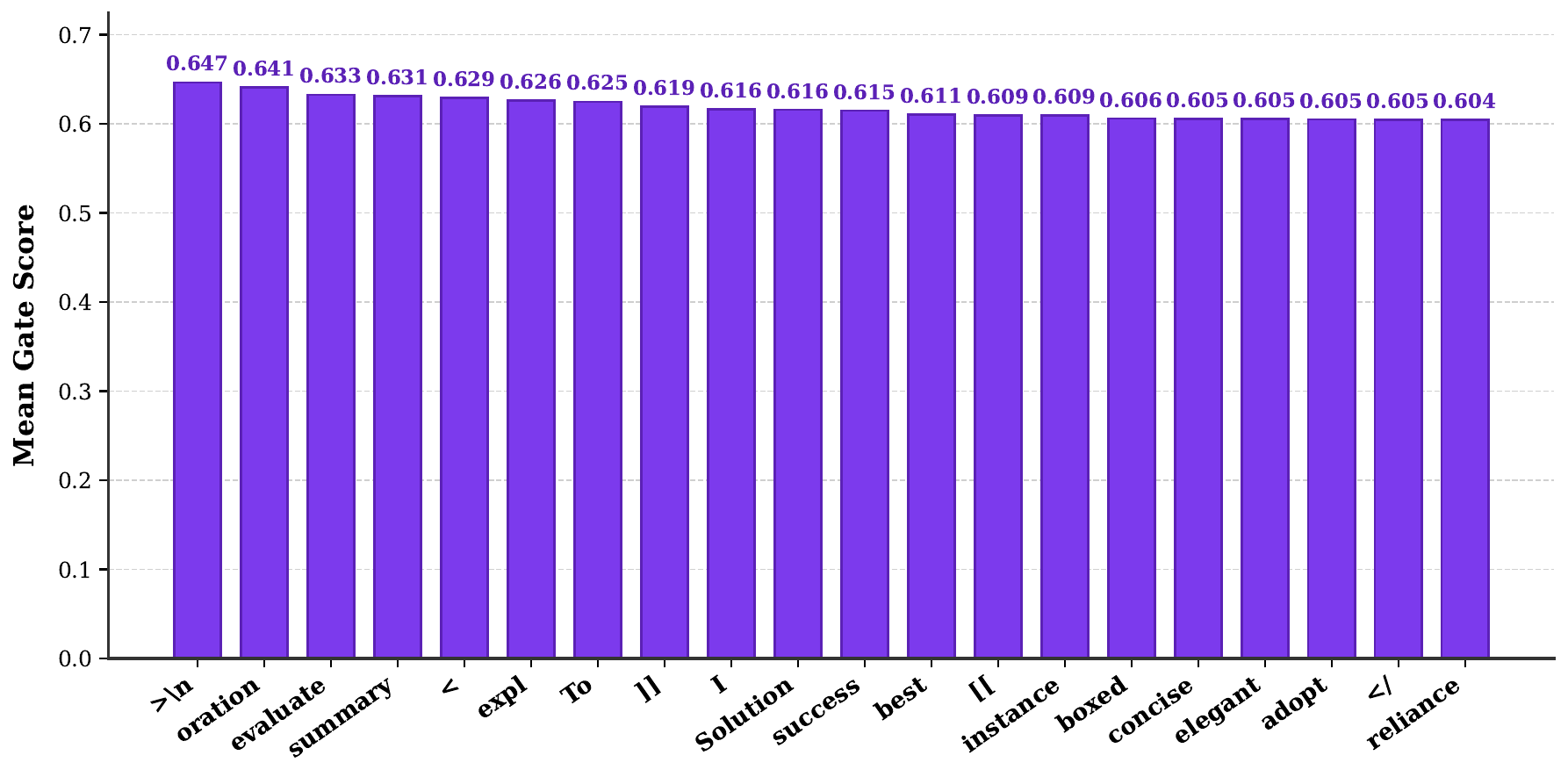}
    \caption{Average gate scores of top-20 tokens ranked by mean gate scores across all threads and samples in AIME 25 by Lattice-1.7B. Tokens related to self-assessment and cross-thread evaluation dominate the rankings, indicating that the model increases cross-thread attention during critical reasoning steps.}
    \label{fig:topk_by_token}
\end{figure}

In \cref{sec:qualitative_analysis}, we show an example thread and visualize the gate scores assigned to each token, concluding that higher gate scores correspond to more critical reasoning steps. Here, we provide additional statistics to support this observation. As shown in \cref{fig:topk_by_token}, we compute the average gate scores of Lattice-1.7B across all threads and samples in AIME 25, and rank the top-20 tokens by their mean gate scores. The results reveal a clear pattern: tokens associated with \textbf{self-assessment and cross-thread evaluation} dominate the rankings. Notably, subword tokens of ``exploration'' (\texttt{expl}, \texttt{oration}) and evaluation-related tokens (\texttt{summary}, \texttt{evaluate}, \texttt{Solution}) rank highly, indicating that the model increases cross-thread attention during the exploration phase. Furthermore, self-selection tags (\textsc{success}, \textsc{best}) also exhibit elevated gate scores, substantiating that the model actively references peer threads when making final judgments. This quantitative evidence supports our qualitative finding that LACE learns to selectively engage cross-thread communication at semantically meaningful positions, particularly during exploration and self-assessment stages. 

\begin{figure}[t]
    \centering
    \includegraphics[width=0.85\linewidth]{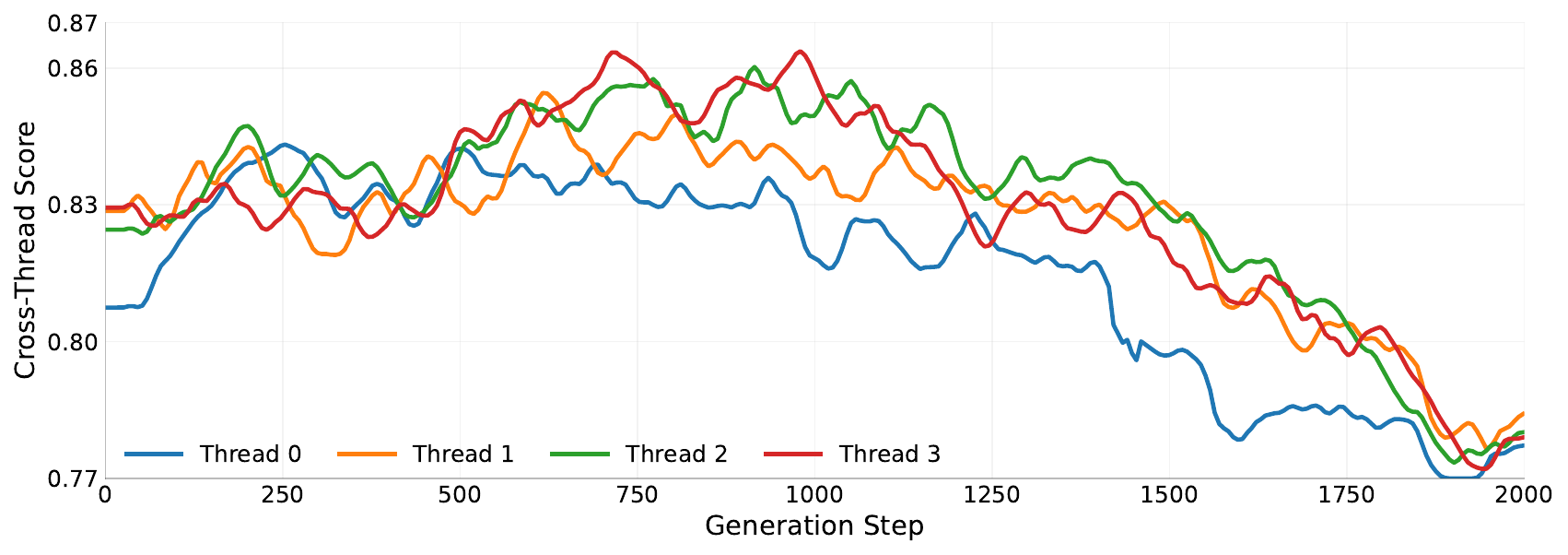}
    \caption{\textbf{Gate-score dynamics over decoding steps.} Information flow in the lattice peaks during the core reasoning phase before tapering off near the final answer, consistent with a diverge-then-converge pattern of collaborative reasoning.}
    \label{fig:gate_score}
\end{figure}

\FloatBarrier

\section{Per-Layer Synchronization Analysis}
\label{appendix:layer_sync}

To further characterize where cross-thread collaboration is concentrated, we measure layer-wise synchronization under teacher forcing on AIME25. We define synchronization as the product of the gate value and the cross-thread attention ratio, i.e., $\text{Sync}=\text{Gate}\times r_{\text{cross}}$, which captures effective cross-thread information flow. \cref{tab:layer_sync} shows that synchronization is concentrated in middle-to-late layers, with Layer 26 consistently serving as the strongest synchronization point across both model scales. RL also strengthens synchronization across nearly all layers, though the largest gains appear at different depths for 1.7B and 4B, suggesting scale-dependent collaboration patterns.

\begin{table}[htbp]
\centering
\scriptsize
\setlength{\tabcolsep}{3pt}
\renewcommand{\arraystretch}{0.95}
\caption{\textbf{Per-layer synchronization analysis on AIME25.} Sync denotes effective cross-thread information flow, computed as gate $\times$ cross-thread ratio. Green row shading highlights RL synchronization profiles, and green cells in $\Delta$ sync mark layer-wise gains after RL.}
\label{tab:layer_sync}
\resizebox{\textwidth}{!}{%
\begin{tabular}{l ccccccccccc}
	\toprule
& \textbf{L10} & \textbf{L12} & \textbf{L14} & \textbf{L16} & \textbf{L18} & \textbf{L20} & \textbf{L22} & \textbf{L24} & \textbf{L26} & \textbf{L28} & \textbf{L30} \\
\midrule
    \textbf{1.7B SFT} sync & 0.232 & 0.185 & 0.178 & 0.194 & 0.235 & 0.251 & 0.231 & 0.246 & \textbf{0.284} & -- & -- \\
    \textbf{1.7B RL} sync & \gainrowcell{0.263} & \gainrowcell{0.228} & \gainrowcell{0.239} & \gainrowcell{0.299} & \gainrowcell{\textbf{0.300}} & \gainrowcell{0.288} & \gainrowcell{0.276} & \gainrowcell{0.280} & \gainrowcell{0.290} & \gainrowcell{--} & \gainrowcell{--} \\
    	\textbf{1.7B} $\Delta$ sync & \gaincell{+0.031} & \gaincell{+0.043} & \gaincell{+0.061} & \gainmaxcell{+0.105} & \gaincell{+0.065} & \gaincell{+0.036} & \gaincell{+0.044} & \gaincell{+0.034} & \gaincell{+0.006} & -- & -- \\
\midrule
    \textbf{4B SFT} sync & 0.210 & 0.189 & 0.160 & 0.168 & 0.149 & 0.162 & 0.161 & 0.149 & \textbf{0.215} & 0.193 & 0.175 \\
    \textbf{4B RL} sync & \gainrowcell{0.213} & \gainrowcell{0.194} & \gainrowcell{0.167} & \gainrowcell{0.176} & \gainrowcell{0.157} & \gainrowcell{0.171} & \gainrowcell{0.171} & \gainrowcell{0.162} & \gainrowcell{\textbf{0.224}} & \gainrowcell{0.207} & \gainrowcell{0.195} \\
    	\textbf{4B} $\Delta$ sync & \gaincell{+0.003} & \gaincell{+0.005} & \gaincell{+0.007} & \gaincell{+0.007} & \gaincell{+0.009} & \gaincell{+0.009} & \gaincell{+0.011} & \gaincell{+0.013} & \gaincell{+0.008} & \gaincell{+0.014} & \gainmaxcell{+0.020} \\
\bottomrule
\end{tabular}%
}
\end{table}

\FloatBarrier

\section{Emergent Cross-Thread Collaboration Examples}
\begin{figure}[!t]
  \centering
  \includegraphics[width=0.95\linewidth, trim=0 0 1.5cm 0, clip]{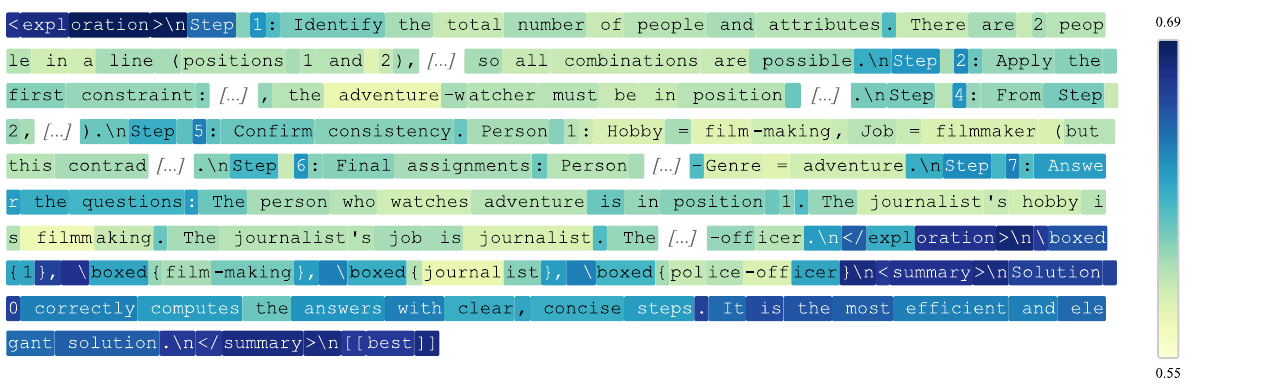}
  \caption{A detailed token-level view of cross-thread intensity on a selected \textsc{best} thread from Lattice-1.7B on an AIME 25 problem. Darker shading indicates stronger cross-thread interaction, with the strongest signals occurring near exploration and self-assessment tokens.}
  \label{fig:full_generation_example}
\end{figure}

By inspecting the generated reasoning traces from \shortname models, we observe several emergent behaviors that suggest enhanced collaborative reasoning capabilities. \cref{fig:full_generation_example} provides a detailed token-level view of cross-thread intensity on a selected \textsc{best} thread from Lattice-1.7B on an AIME 25 problem.

As shown in \cref{fig:full_generation_example}, cross-thread attention peaks around semantically meaningful positions, especially during exploration and self-assessment. This complements the main-text four-thread visualization in \cref{fig:example_generation}, showing more clearly how the selected thread absorbs peer information at the token level.

More interestingly, the full thread summaries in \cref{fig:thread_summaries} reveal an \textbf{emergent early-stopping phenomenon}: once Thread 1 arrives at the correct answer via an efficient approach, Threads 2 and 4 recognize this and conclude that their solutions ``follow a similar approach to the reference but with slightly different phrasing.'' Rather than redundantly completing their own full derivations, these threads \textbf{defer to the superior solution} and label themselves as \textsc{success}. This behavior demonstrates that \shortname enables threads to learn from each other's progress in real-time and avoid redundant computation.

\definecolor{thread1}{RGB}{80,73,202}
\definecolor{thread1bg}{RGB}{237,236,252}
\definecolor{thread2}{RGB}{16,185,129}
\definecolor{thread2bg}{RGB}{232,250,245}
\definecolor{thread3}{RGB}{245,158,11}
\definecolor{thread3bg}{RGB}{254,247,235}
\definecolor{thread4}{RGB}{236,72,153}
\definecolor{thread4bg}{RGB}{253,237,246}

\begin{figure*}[h]
\centering
\begin{minipage}{0.48\textwidth}
\begin{tcolorbox}[colback=thread1bg, colframe=thread1, title=\textbf{Thread 1 Summary}, fonttitle=\bfseries, left=2mm, right=2mm]
\small
Solution 0 correctly finds the bases 21 and 49 by analyzing the division and checking divisors of 56. It is the most direct and efficient solution.

\vspace{0.3em}
\textbf{\textsc{best}}
\end{tcolorbox}
\end{minipage}
\hfill
\begin{minipage}{0.48\textwidth}
\begin{tcolorbox}[colback=thread2bg, colframe=thread2, title=\textbf{Thread 2 Summary}, fonttitle=\bfseries, left=2mm, right=2mm]
\small
This solution follows a similar approach to the reference but with slightly different phrasing. It is correct and efficient.

\vspace{0.3em}
\textbf{\textsc{success}}
\end{tcolorbox}
\end{minipage}
\end{figure*}

\begin{figure*}[h]
\centering
\begin{minipage}{0.48\textwidth}
\begin{tcolorbox}[colback=thread3bg, colframe=thread3, title=\textbf{Thread 3 Summary}, fonttitle=\bfseries, left=2mm, right=2mm]
\small
This solution uses modular arithmetic and the divisibility property to derive the same result as the reference method. It is concise and efficient.

\vspace{0.3em}
\textbf{\textsc{success}}
\end{tcolorbox}
\end{minipage}
\hfill
\begin{minipage}{0.48\textwidth}
\begin{tcolorbox}[colback=thread4bg, colframe=thread4, title=\textbf{Thread 4 Summary}, fonttitle=\bfseries, left=2mm, right=2mm]
\small
Solution 3 follows a similar algebraic approach as Solution 2 but with slightly more verbose steps. It is correct but less efficient than Solution 2.

\vspace{0.3em}
\textbf{\textsc{success}}
\end{tcolorbox}
\end{minipage}
\caption{Self-assessment summaries from four parallel threads on an AIME 25 problem. Thread 1 correctly identifies itself as the best solution, while other threads acknowledge their relative positions.}
\label{fig:thread_summaries}
\end{figure*}

\FloatBarrier

\section{Additional Experimental Analyses}
\label{appendix:additional_experiments}

\subsection{Computational Overhead Micro-Benchmark}
\label{appendix:overhead}

\begin{table}[!t]
\centering
\scriptsize
\setlength{\tabcolsep}{2pt}
\renewcommand{\arraystretch}{0.95}
\caption{\textbf{Thread-scaling micro-benchmark on Lattice-1.7B.} Memory is reported in MB. Red shading marks overhead increases.}
\label{tab:overhead_17b}
\resizebox{\textwidth}{!}{%
\begin{tabular}{c ccccc ccccc cccc}
  \toprule
& \multicolumn{5}{c}{\textbf{Vanilla}} & \multicolumn{5}{c}{\textbf{Lattice}} & \multicolumn{4}{c}{\textbf{Overhead}} \\
\cmidrule(lr){2-6} \cmidrule(lr){7-11} \cmidrule(lr){12-15}
  \textbf{N} & \textbf{ms/step} & \textbf{TPS} & \textbf{TTFT (ms)} & \textbf{Mem} & \textbf{FLOPs} & \textbf{ms/step} & \textbf{TPS} & \textbf{TTFT (ms)} & \textbf{Mem} & \textbf{FLOPs} & \textbf{ms/step} & \textbf{TTFT} & \textbf{Mem} & \textbf{FLOPs} \\
\midrule
1 & 20.8 & 48.1 & 18.1 & 3,911 & $1.90 \times 10^{11}$ & -- & -- & -- & -- & -- & -- & -- & -- & -- \\
4 & 20.5 & 195.4 & 18.7 & 4,003 & $7.60 \times 10^{11}$ & 28.4 & 140.8 & 25.8 & 4,069 & $7.69 \times 10^{11}$ & \costcell{+38.5\%} & \costcell{+38.0\%} & \costcell{+1.6\%} & \costcell{+1.22\%} \\
8 & 20.5 & 391.2 & 18.6 & 7,993 & $1.52 \times 10^{12}$ & 28.7 & 279.1 & 26.0 & 8,128 & $1.54 \times 10^{12}$ & \costcell{+40.0\%} & \costmaxcell{39.8\%} & \costcell{+1.7\%} & \costcell{+1.22\%} \\
16 & 20.4 & 783.6 & 25.8 & 8,217 & $3.04 \times 10^{12}$ & 31.0 & 515.5 & 29.8 & 8,416 & $3.08 \times 10^{12}$ & \costcell{+51.9\%} & \costcell{+15.5\%} & \costcell{+2.4\%} & \costcell{+1.22\%} \\
32 & 21.6 & 1,480 & 33.8 & 8,673 & $6.08 \times 10^{12}$ & 33.2 & 964.8 & 38.2 & 8,997 & $6.15 \times 10^{12}$ & \costcell{+53.7\%} & \costcell{+13.0\%} & \costcell{+3.7\%} & \costcell{+1.23\%} \\
64 & 22.5 & 2,845 & 51.9 & 9,585 & $1.22 \times 10^{13}$ & 35.2 & 1,817 & 57.0 & 10,161 & $1.23 \times 10^{13}$ & \costcell{+56.4\%} & \costcell{+9.8\%} & \costcell{+6.0\%} & \costcell{+1.25\%} \\
128 & 25.8 & 4,968 & 91.5 & 11,181 & $2.43 \times 10^{13}$ & 41.0 & 3,124 & 100.1 & 12,278 & $2.46 \times 10^{13}$ & \costmaxcell{58.9\%} & \costcell{+9.4\%} & \costmaxcell{9.8\%} & \costmaxcell{1.29\%} \\
\bottomrule
\end{tabular}%
}
\end{table}

\begin{table}[!t]
\centering
\scriptsize
\setlength{\tabcolsep}{2pt}
\renewcommand{\arraystretch}{0.95}
\caption{\textbf{Thread-scaling micro-benchmark on Lattice-4B.} Memory is reported in MB. Red shading marks overhead increases.}
\label{tab:overhead_4b}
\resizebox{\textwidth}{!}{%
\begin{tabular}{c ccccc ccccc cccc}
  \toprule
& \multicolumn{5}{c}{\textbf{Vanilla}} & \multicolumn{5}{c}{\textbf{Lattice}} & \multicolumn{4}{c}{\textbf{Overhead}} \\
\cmidrule(lr){2-6} \cmidrule(lr){7-11} \cmidrule(lr){12-15}
  \textbf{N} & \textbf{ms/step} & \textbf{TPS} & \textbf{TTFT (ms)} & \textbf{Mem} & \textbf{FLOPs} & \textbf{ms/step} & \textbf{TPS} & \textbf{TTFT (ms)} & \textbf{Mem} & \textbf{FLOPs} & \textbf{ms/step} & \textbf{TTFT} & \textbf{Mem} & \textbf{FLOPs} \\
\midrule
1 & 26.2 & 38.5 & 23.1 & 7,767 & $4.42 \times 10^{11}$ & -- & -- & -- & -- & -- & -- & -- & -- & -- \\
4 & 27.6 & 144.8 & 23.9 & 7,874 & $1.77 \times 10^{12}$ & 36.2 & 110.6 & 33.1 & 8,717 & $1.77 \times 10^{12}$ & \costcell{+31.2\%} & \costmaxcell{38.5\%} & \costcell{+10.7\%} & \costcell{+0.35\%} \\
8 & 26.3 & 304.6 & 27.3 & 15,745 & $3.54 \times 10^{12}$ & 36.2 & 221.2 & 33.5 & 17,417 & $3.55 \times 10^{12}$ & \costcell{+37.6\%} & \costcell{+22.7\%} & \costcell{+10.6\%} & \costcell{+0.36\%} \\
16 & 26.6 & 602.7 & 36.4 & 16,007 & $7.07 \times 10^{12}$ & 39.6 & 403.6 & 41.8 & 17,766 & $7.10 \times 10^{12}$ & \costcell{+48.9\%} & \costcell{+14.8\%} & \costcell{+11.0\%} & \costcell{+0.36\%} \\
32 & 28.8 & 1,112 & 61.5 & 16,551 & $1.41 \times 10^{13}$ & 42.1 & 759.7 & 66.6 & 18,473 & $1.42 \times 10^{13}$ & \costcell{+46.2\%} & \costcell{+8.3\%} & \costcell{+11.6\%} & \costcell{+0.36\%} \\
64 & 29.6 & 2,161 & 108.5 & 17,641 & $2.83 \times 10^{13}$ & 45.4 & 1,409 & 115.5 & 19,887 & $2.84 \times 10^{13}$ & \costcell{+53.4\%} & \costcell{+6.5\%} & \costcell{+12.7\%} & \costcell{+0.36\%} \\
128 & 33.3 & 3,842 & 201.4 & 19,594 & $5.66 \times 10^{13}$ & 65.4 & 1,958 & 211.4 & 22,506 & $5.68 \times 10^{13}$ & \costmaxcell{96.4\%} & \costcell{+5.0\%} & \costmaxcell{14.9\%} & \costmaxcell{0.37\%} \\
\bottomrule
\end{tabular}%
}
\end{table}

We benchmark thread scaling on a single RTX PRO 6000 Blackwell 97GB GPU using bfloat16 precision, generation length 100, and context length 50. We compare vanilla independent sampling against LACE-enabled decoding across thread counts from $N=1$ to $N=128$. The results in \cref{tab:overhead_17b,tab:overhead_4b} show that LACE adds negligible FLOPs overhead (below $1.3\%$), while the dominant cost is step latency, indicating that the extra overhead is primarily memory-bandwidth bound rather than compute bound. Memory overhead remains modest, and TTFT overhead decreases at larger thread counts. Notably, even 128 threads fit on a single GPU, requiring only about 12.3GB for Lattice-1.7B and 22.5GB for Lattice-4B.

\FloatBarrier

\section{Limitations}
\label{appendix:limitations}

The main limitation of LACE is the additional cross-thread communication cost introduced by lattice attention. This overhead is expected because multiple reasoning threads exchange intermediate states during generation, whereas independent sampling keeps the threads isolated. However, the routing path is lightweight and our micro-benchmarks in \cref{tab:overhead_17b,tab:overhead_4b} show that the cost is tolerable in practice: FLOPs overhead remains below $1.3\%$, memory overhead is modest, and the dominant latency cost is primarily memory-bandwidth bound rather than compute-bound.

\FloatBarrier

\section{Broader Impacts}
\label{appendix:broader_impacts}

LACE may improve the reliability and efficiency of reasoning models by allowing multiple solution paths to coordinate during generation. Potential benefits include stronger mathematical reasoning, planning, and interactive problem solving. At the same time, stronger reasoning and agentic capabilities may also be dual-use, including possible misuse for automated problem solving in harmful settings. Responsible deployment should therefore follow standard model safety practices, including careful evaluation, access controls when appropriate, and monitoring for misuse.

\FloatBarrier

\subsection{Diversity Reward Sensitivity}
\label{appendix:diversity_sensitivity}
\begin{table}[htbp]
\centering
\small
\caption{\textbf{Relative roles of accuracy reward and diversity reward.} Due to limited computational resources, we report results on the subset training split using Lattice-1.7B. All numbers are percentages.}
\label{tab:diversity_sensitivity}
\begin{tabular}{l ccc ccc}
  \toprule
  & \multicolumn{3}{c}{\textbf{Acc}} & \multicolumn{3}{c}{\textbf{Pass@4}} \\
  \cmidrule(lr){2-4} \cmidrule(lr){5-7}
  	\textbf{Dataset} & \textbf{Div=0} & \textbf{Div=1.0} & \textbf{Div=2.0} & \textbf{Div=0} & \textbf{Div=1.0} & \textbf{Div=2.0} \\
\midrule
AIME24 & 3.3 & \textbf{13.3} & 6.7 & 13.3 & 13.3 & \textbf{16.7} \\
DAPO-Test100 & 23.0 & \textbf{30.0} & 26.0 & 51.0 & \textbf{58.0} & 53.0 \\
\bottomrule
\end{tabular}
\end{table}
We conduct this study on the subset training split using the 1.7B model. A useful intuition is that the diversity reward acts like a training-time temperature, reshaping the model's path distribution during training. As shown in \cref{tab:diversity_sensitivity}, the optimal diversity level depends on the task: on more concentrated tasks such as DAPO-Test100, excessive diversity can over-spread search and hurt accuracy; on harder tasks such as AIME24, stronger diversity can improve coverage (Pass@4), but the rarer successful paths then become harder to stably self-select. Overall, the baseline weight provides the best balance between path coverage and focused best-path selection.

\FloatBarrier

\subsection{Agent Benchmarks}
\label{appendix:textworld_generalization}

\begin{table}[htbp]
\centering
\small
\caption{\textbf{Results on TextWorldCookAgent.} We report Best Score, Mean Score, and Win Rate for Lattice-1.7B and matched baselines.}
\label{tab:textworld_generalization}
\begin{tabular}{l c c c}
  \hline
  {\bfseries Model} & {\bfseries Best Score} & {\bfseries Mean Score} & {\bfseries Win Rate} \\
\midrule
Qwen3-1.7B & 61.0 & 33.6 & 37.0 \\
Qwen3-1.7B + Single-Thread GRPO & 63.5 & 39.2 & 37.0 \\
Qwen3-1.7B + 4-Thread GRPO (Isolated Parallel) & 60.9 & 36.4 & 34.0 \\
{\bfseries Lattice-1.7B + 4-Thread GRPO (Ours)} & {\bfseries 67.6} & 38.1 & {\bfseries 45.0} \\
  \hline
\end{tabular}
\end{table}

To test whether the same collaborative design is also effective in interactive agent settings, we additionally train and evaluate Lattice-1.7B on TextWorldCookAgent from the TALES benchmark suite built on TextWorld~\citep{cui2025tales,cote18textworld}, an interactive task requiring longer-horizon exploration and intermediate feedback. As shown in \cref{tab:textworld_generalization}, LACE achieves the highest Best Score and Win Rate, outperforming both single-thread RL and isolated multi-thread RL. The Mean Score remains slightly below the single-thread GRPO baseline, suggesting that the gains are strongest in peak performance and head-to-head competitiveness rather than uniform average returns. Overall, these results provide initial evidence that the collaborative mechanism in LACE remains beneficial in agent tasks as well, while broader evaluation on more open-ended domains remains important future work.

\FloatBarrier

\subsection{Thread Scaling and Extended Efficiency Results}
\label{appendix:thread_scaling}

\begin{table}[htbp]
\centering
\small
\caption{\textbf{Inference-time thread scaling on TextWorldCookAgent.} Models are trained with 4 threads and evaluated with larger decoding thread counts without retraining.}
\label{tab:thread_scaling_agent}
\begin{tabular}{l c c c}
  	\toprule
  	\textbf{Metric} & \textbf{4T} & \textbf{8T} & \textbf{16T} \\
  \midrule
  Mean Score & 38.1 & 40.6 & 40.6 \\
  Best Score & 67.6 & 75.3 & 83.1 \\
  Win Rate & 45.0 & 57.0 & 68.0 \\
  \bottomrule
\end{tabular}
\end{table}

\begin{table}[htbp]
\centering
\small
\caption{\textbf{Training-time thread scaling on LiveBench.} We compare 4-thread and 8-thread Lattice-1.7B models trained on the subset split.}
\label{tab:thread_scaling_livebench}
\begin{tabular}{l c c}
  	\toprule
  	\textbf{Metric} & \textbf{4T train} & \textbf{8T train} \\
  \midrule
  Accuracy & 6.2 & 12.3 \\
  Latency / query (s) & 67.9 & 21.2 \\
  Tokens / query & 6,434 & 2,107 \\
  \bottomrule
\end{tabular}
\end{table}

\begin{table}[htbp]
\centering
\small
\caption{\textbf{Preliminary scaling to 8B on the subset split.} We report best-thread accuracy and four-sample mean accuracy.}
\label{tab:model_scale_preliminary}
\begin{tabular}{l c c}
  	\toprule
  	\textbf{Metric} & \textbf{Lattice-8B} & \textbf{Qwen3-8B + Single SFT + GRPO} \\
  \midrule
  AIME24 Acc & 16.7 & 10.8 \\
  AIME24 Mean@4 & 27.5 & 10.8 \\
  DAPO-Test100 Acc & 51.0 & 51.7 \\
  DAPO-Test100 Mean@4 & 61.5 & 51.7 \\
  \bottomrule
\end{tabular}
\end{table}

The results in \cref{tab:thread_scaling_agent,tab:thread_scaling_livebench,tab:model_scale_preliminary} suggest that the collaborative mechanism extrapolates beyond the training-time thread count at inference time, and that increasing thread count during training can improve both the probability of early successful self-selection and end-to-end efficiency. The 8B-scale results in \cref{tab:model_scale_preliminary} remain preliminary, but already indicate that cross-thread collaboration transfers to larger models before self-assessment behavior is fully stabilized.

\FloatBarrier

\subsection{Additional Baseline Comparisons}
\label{appendix:baseline_ext}

\begin{table}[htbp]
\centering
\small
\caption{\textbf{Additional selection-based baselines on AIME24 and AIME25.} We compare Lattice-4B against judge-based selection~\citep{zheng2023judging} and sequential refinement~\citep{madaan2023self} baselines.}
\label{tab:additional_baselines}
\resizebox{\textwidth}{!}{%
\begin{tabular}{l ccc ccc}
  	\toprule
  & \multicolumn{3}{c}{\textbf{AIME24}} & \multicolumn{3}{c}{\textbf{AIME25}} \\
  \cmidrule(lr){2-4} \cmidrule(lr){5-7}
  	\textbf{Method} & \textbf{Acc} & \textbf{Latency/q} & \textbf{Tokens/q} & \textbf{Acc} & \textbf{Latency/q} & \textbf{Tokens/q} \\
  \midrule
  LACE & 20.0 & 61.2s & 7,120 & 20.0 & 53.0s & 7,012 \\
  Gen-Judge (self) & 0.0 & 97.8s & 10,573 & 0.0 & 105.3s & 10,886 \\
  Gen-Judge (oracle-4o / 4.1) & 20.0 / 16.7 & 86.7s / 98.6s & 9,069 / 10,041 & 13.3 / 20.0 & 94.5s / 106.8s & 9,499 / 10,237 \\
  Seq-Ref & 16.7 & $\sim$ 245s & 15,446 & 13.3 & $\sim$ 296s & 14,369 \\
  \bottomrule
\end{tabular}%
}
\end{table}

As shown in \cref{tab:additional_baselines}, judge-based post-hoc selection~\citep{zheng2023judging} can be competitive only when paired with a stronger external judge, but it requires additional decoding and verification cost. Sequential refinement~\citep{madaan2023self} similarly incurs substantial extra decoding length. In contrast, LACE performs self-selection in situ during generation, retaining lower end-to-end latency and token usage while remaining competitive in final accuracy.

\FloatBarrier

\subsection{Self-Assessment and Identity Ablations}
\label{appendix:sa_identity_ablation}

\begin{table}[htbp]
\centering
\small
\caption{\textbf{Ablation of self-assessment supervision.} The w/o-SA setting removes self-assessment tags at inference time and falls back to mean@4 or voting@4 aggregation.}
\label{tab:sa_ablation}
\begin{tabular}{l c c c}
  	\toprule
  	\textbf{Benchmark} & \textbf{w/ SA } & \textbf{w/o SA (mean@4)} & \textbf{w/o SA (voting@4)} \\
  \midrule
  AIME24 & 20.0 & 7.5 & 6.7 \\
  AIME25 & 16.7 & 5.8 & 6.7 \\
  \bottomrule
\end{tabular}
\end{table}

\begin{table}[htbp]
\centering
\small
\caption{\textbf{Ablation of path identity encoding on TextWorldCookAgent.} Prompt path identifiers provide the strongest signal, while 3D RoPE remains complementary.}
\label{tab:path_id_ablation}
\begin{tabular}{c c c}
  	\toprule
  	\textbf{Prompt Path ID} & \textbf{RoPE Path ID} & \textbf{Best Score} \\
  \midrule
  $\checkmark$ & $\checkmark$ & 67.6 \\
  $\checkmark$ & $\times$ & 66.7 \\
  $\times$ & $\checkmark$ & 64.5 \\
  \bottomrule
\end{tabular}
\end{table}

The results in \cref{tab:sa_ablation,tab:path_id_ablation} indicate that self-assessment is crucial for identifying the best thread among explored candidates, while thread identity is not carried by RoPE alone: prompt-level identifiers already provide a strong signal, and combining both cues yields the strongest performance.

\FloatBarrier

\subsection{Training Data Statistics}
\label{appendix:data_stats}

\begin{table}[htbp]
\centering
\small
\caption{\textbf{Training and evaluation data statistics used in this work.}}
\label{tab:data_stats}
\begin{tabular}{l c c}
  	\toprule
  	\textbf{Training Set} & \textbf{Stage} & \textbf{Questions} \\
  \midrule
  Lattice-1.7B & SFT & 800 \\
  Lattice-1.7B & RL & 1,959 \\
  Lattice-4B & SFT & 2,409 \\
  Lattice-4B & RL & 6,474 \\
  Lattice-1.7B-Subset & SFT & 576 \\
  Lattice-1.7B-Subset & RL & 960 \\
  DAPO-Test100 & Held-out Test & 100 \\
  \bottomrule
\end{tabular}
\end{table}

All SFT splits contain at least four candidate chains-of-thought per problem. We use the subset split only for the additional scaling and ablation experiments reported above.

\FloatBarrier

\section{Implementation Details}
\label{appendix:implementation}
\subsection{Model Training}
We implement LACE in Qwen3 models with 1.7B and 4B parameters. The 1.7B model is initialized from Qwen3-1.7B with disabled thinking for efficient training, while the 4B model is initialized from Qwen3-4B-Instruct-2507. Both models undergo SFT and RL training as described in \cref{sec:training}. During the initial SFT steps, we freeze the backbone and update only the lattice attention parameters before unfreezing the full model for the remaining SFT steps. They are trained on the DAPO dataset processed through the data curation pipeline outlined in \cref{sec:data}. For SFT, the learning rate is set to $1 \times 10^{-5}$ with a cosine scheduler. For RL training, the learning rate is reduced to $1 \times 10^{-6}$. We utilize the AdamW optimizer with a weight decay of 0.01 and betas of (0.9, 0.95). The 1.7B model is evaluated after 200 RL steps for AIME 24 and AIME 25, while the 4B model is evaluated after 320 RL steps. For LiveBench, due to the larger domain gap compared to AIME, we use a later checkpoint for the 1.7B model at 350 RL steps, while the 4B model remains at 320 RL steps.

\section{Data Curation Pipeline Prompts}
\label{appendix:data_curation}
As mentioned in \cref{sec:data}, our data, sourced from the DAPO dataset, undergoes a rigorous curation process to ensure high-quality training data. We use an iterative generation process with a solution cache to enhance reasoning diversity. The inference model is Qwen3-235B-A22B. For diverse generation, we employ the prompt template in \cref{fig:diverse_prompt}:

\definecolor{mainpurple}{RGB}{80,73,202}
\definecolor{lightpurple}{RGB}{237,236,252}

\begin{figure}[h]
\begin{tcolorbox}[colback=lightpurple, colframe=mainpurple, title=\textbf{Diverse Solution Generation Prompt}, fonttitle=\bfseries, left=2mm, right=2mm]
\small
\textbf{System:} You are a creative mathematician who excels at finding alternative problem-solving strategies. Your goal is to solve the given problem using a method that is fundamentally different from the provided reference approach. You must maintain rigorous logic and mathematical precision.

\vspace{0.5em}
\textbf{Problem:} \texttt{\{Problem\_Statement\}}

\vspace{0.5em}
\textbf{Reference Method (DO NOT USE):}\\
\texttt{\{solution\_summary\}}

\vspace{0.5em}
\textbf{Task:}\\
Solve the problem above using a distinct method.\\
1. \textbf{Constraint}: You are strictly FORBIDDEN from using the reasoning logic described in the ``Reference Method''.\\
2. \textbf{Goal}: Find a distinct path (e.g., if Reference uses Algebra, try Geometry or Number Theory).\\
3. \textbf{Format}: Put your final answer within \texttt{\textbackslash boxed\{\}}.

\vspace{0.5em}
\hrule
\vspace{0.5em}
\textbf{Example of Distinct Methods:}

\textit{Problem:} Find the distance between point A(1,1) and B(4,5).\\
\textit{Reference Method:} Used the Pythagorean theorem explicitly by constructing a right triangle with legs 3 and 4.

\textit{Your New Solution:}\\
We recognize this as a distance in the Cartesian plane. We apply the standard Cartesian distance formula directly:
$d = \sqrt{(x_2 - x_1)^2 + (y_2 - y_1)^2}$\\
Substituting the coordinates:
$d = \sqrt{(4-1)^2 + (5-1)^2} = \sqrt{3^2 + 4^2} = \sqrt{25} = 5.$\\
The distance is 5. $\boxed{5}$

\vspace{0.5em}
\hrule
\vspace{0.5em}
Now, strictly following the constraints, reasoning step by step, solve the original problem and put your final answer within \texttt{\textbackslash boxed\{\}}:
\end{tcolorbox}
\caption{Prompt template for diverse solution generation. The \texttt{\{solution\_summary\}} placeholder is populated with cached summaries of previously generated solutions to encourage distinct reasoning paths.}
\label{fig:diverse_prompt}
\end{figure}

For solution summary, we employ the template in \cref{fig:summary_prompt}:

\begin{figure}[h]
\begin{tcolorbox}[colback=lightpurple, colframe=mainpurple, title=\textbf{Solution Summary Prompt}, fonttitle=\bfseries, left=2mm, right=2mm]
\small
\textbf{System:} You are an expert at abstracting mathematical problem-solving strategies. Your goal is to describe the high-level methodology used in a solution without revealing specific numbers or calculations.

\vspace{0.5em}
\textbf{Problem:} \texttt{\{Problem\_Statement\}}\\
\textbf{Solution:} \texttt{\{Cot\_Solution\}}

\vspace{0.5em}
\textbf{Task:}\\
Summarize the \textit{methodology} used in the solution above into a high-level abstract.\\
1. \textbf{Constraint}: Maximum 3 sentences. Keep it under 60 words.\\
2. \textbf{CRITICAL}: Do NOT include specific numbers from the calculation (e.g., replace ``15'' with ``the total count'', ``$x^2$'' with ``the quadratic term'').\\
3. Focus on the logical steps (e.g., ``Use the Pythagorean theorem,'' ``Apply substitution,'' ``Analyze modulo residues'').\\
4. Do not reveal the final answer.

\vspace{0.5em}
Format your output strictly as:\\
\texttt{\$\$\$summary\$\$\$}\\
\texttt{[Your abstract summary here]}\\
\texttt{\$\$\$summary\$\$\$}

\vspace{0.5em}
\hrule
\vspace{0.5em}
\textbf{Example:}

\textit{Problem:} Find $x$ if $2x + 6 = 16$.\\
\textit{Solution:} Subtract 6 from both sides to get $2x = 10$. Then divide by 2 to get $x = 5$.

\texttt{\$\$\$summary\$\$\$}\\
The solution isolates the variable by first performing subtraction to remove the constant term and then dividing by the coefficient of the variable.\\
\texttt{\$\$\$summary\$\$\$}

\vspace{0.5em}
\hrule
\vspace{0.5em}
Now, generate the summary for the provided solution:
\end{tcolorbox}
\caption{Prompt template for solution summarization. The generated summaries are cached and used to guide diverse solution generation in subsequent iterations. \textbf{Original prompt is very long and here is a shortened version.}}
\label{fig:summary_prompt}
\end{figure}

For step decomposition, we use the prompt template in \cref{fig:step_decomposition_prompt}:

\begin{figure}[h]
\begin{tcolorbox}[colback=lightpurple, colframe=mainpurple, title=\textbf{Step Decomposition Prompt}, fonttitle=\bfseries, left=2mm, right=2mm]
\small
\textbf{System:} You are an expert Mathematics Reasoning Editor. Your task is to rewrite a verbose, potentially repetitive, or erroneous ``Draft CoT'' (Chain-of-Thought) into a concise, structured, and correct ``Standard CoT''.

\vspace{0.5em}
\textbf{Problem:} \texttt{\{question\}}\\
\textbf{Standard Answer:} \texttt{\{ground\_truth\}}\\
\textbf{Draft CoT:} \texttt{\{draft\_cot\}}\\
\textbf{Final Solution:} \texttt{\{solution\}}

\vspace{0.5em}
\textbf{Rules:}\\
1. \textbf{Length Control}: Ensure the total length is between 1.0 to 3.0 times the length of the ``Reference Solution''. Avoid excessive brevity, but eliminate unnecessary verbosity.\\
2. \textbf{Step-by-Step Structure}: Decompose the reasoning into 4 to 8 distinct steps. Each step should be of roughly uniform length and contain a complete logical deduction.\\
3. \textbf{Content Correction}: If the ``Draft CoT'' contains valid reasoning, preserve and optimize it. If it enters an infinite loop, repetitive pattern, or logical fallacy, truncate immediately and append: ``I have encountered a difficulty and cannot solve this problem.''\\
4. \textbf{Format}: Output strictly in JSON format with \texttt{status} (``solved'' or ``failed'') and \texttt{reasoning\_steps} (list of strings).

\vspace{0.5em}
\hrule
\vspace{0.5em}
\textbf{Example (Successful):}
\begin{verbatim}
{
  "status": "solved",
  "reasoning_steps": [
    "Step 1: Identify the total number of letters...",
    "Step 2: Determine the frequency of each letter...",
    ...
    "Step 8: Conclude that there are 60 distinct ways."
  ]
}
\end{verbatim}

\textbf{Example (Failed):}
\begin{verbatim}
{
  "status": "failed",
  "reasoning_steps": [
    "Step 1: Let the unknown number be x...",
    ...
    "Step 6: ... I have encountered a difficulty 
     and cannot solve this problem."
  ]
}
\end{verbatim}
\end{tcolorbox}
\caption{Prompt template for step decomposition. Verbose or erroneous draft reasoning is rewritten into a concise, structured format with 4--8 steps.}
\label{fig:step_decomposition_prompt}
\end{figure}

For evaluation and labeling of generated solutions, we use the prompt template in \cref{fig:evaluation_prompt}:

\begin{figure}[h]
\begin{tcolorbox}[colback=lightpurple, colframe=mainpurple, title=\textbf{Solution Evaluation Prompt}, fonttitle=\bfseries, left=2mm, right=2mm]
\small
\textbf{System:} You are an expert math reasoning judge. Your task is to evaluate multiple reasoning processes to a math problem and output results in strict JSON format.

\vspace{0.5em}
\textbf{Goal:} Evaluate \texttt{\{num\_items\}} reasoning trajectories to identify the Best Solution, the one demonstrating \textit{Mathematical Elegance}: the perfect balance of Insight and Simplicity.

\vspace{0.5em}
\textbf{Evaluation Criteria:}\\
1. \textbf{Correctness} (Prerequisite): Valid logic leading to correct answer.\\
2. \textbf{Exploration \& Insight}: Complexity of solution must match problem complexity.\\
3. \textbf{Efficiency \& Simplicity}: ``Straight-through'' execution; token economy; no over-engineering or verbose explaining.

\vspace{0.5em}
\textbf{Label Hierarchy:}\\
$\bullet$ \textsc{best}: Elegant, correct, optimal approach, high efficiency. Only ONE solution.\\
$\bullet$ \textsc{success}: Competent, correct but brute force, recovered, or verbose.\\
$\bullet$ \textsc{fail}: Incomplete, incorrect, or invalid logic.

\vspace{0.5em}
\hrule
\vspace{0.5em}
\textbf{Examples:}\\
\textit{Complex Problem (Sum 1 to 100):} Direct formula $n(n+1)/2$ is \textsc{best}; brute force is \textsc{success}.\\
\textit{Simple Problem ($12 \times 15$):} Direct computation is \textsc{best}; algebraic expansion is \textsc{success}.\\
\textit{Verbose Trap ($20+20$):} ``$20+20=40$'' is \textsc{best}; lengthy decomposition is \textsc{success}.

\vspace{0.5em}
\hrule
\vspace{0.5em}
\textbf{Input:}\\
\texttt{Question: \{question\}}\\
\texttt{Ground Truth: \{answer\}}\\
\texttt{Candidate Solutions: \{solutions\_block\}}

\vspace{0.5em}
\textbf{Output Format:}
\begin{verbatim}
{
  "reasoning": "Comparative analysis...",
  "correctness": ["Correct", "Wrong", ...],
  "exploration_rank": [idx0, idx1, ...],
  "efficiency_rank": [idx0, idx1, ...],
  "labels": ["best", "success", ...]
}
\end{verbatim}
\end{tcolorbox}
\caption{Prompt template for solution evaluation. The judge assesses multiple candidate solutions based on correctness, insight, and efficiency, assigning hierarchical labels for training data curation. \textbf{Original prompt is very long and here is a shortened version.}}
\label{fig:evaluation_prompt}
\end{figure}

For evaluation, we follow the recommended setting of Qwen3, with the temperature set to 0.6 and top-p set to 0.9. The inference prompt is shown in \cref{fig:parallel_inference_prompt}:

\begin{figure}[h]
\begin{tcolorbox}[colback=lightpurple, colframe=mainpurple, title=\textbf{Parallel Reasoning Inference Prompt}, fonttitle=\bfseries, left=2mm, right=2mm]
\small
\textbf{System:} You are an expert math problem solver with a parallel reasoning architecture, designed to explore multiple solutions simultaneously. You are \texttt{\{path\_index\}}-th out of \texttt{\{num\_items\}} parallel reasoning paths. Use potential insights from other parallel paths to guide your steps and discover unique strategies for THIS path.

\vspace{0.5em}
\textbf{User:} Given the math problem below, multiple solution paths are being explored in parallel. Your task is to generate the reasoning for solution path \texttt{\{path\_index\}}.

First, reason step by step, considering insights from the other parallel paths. Then, give the solution. Put your final answer within \texttt{\textbackslash boxed\{\}}.

After giving the solution, summarize your solution with comparison with other paths within 3 sentences, enclosed in \texttt{<summary>...</summary>}.

Finally, label your solution with one of the following:\\
$\bullet$ \textsc{best} if you believe this path is the most elegant and efficient among all parallel paths. ONLY one path should use this label.\\
$\bullet$ \textsc{success} if this path leads to the correct answer but is not the best among all parallel paths.\\
$\bullet$ \textsc{fail} if this path does not lead to the correct answer.

\vspace{0.5em}
\hrule
\vspace{0.5em}
\textbf{Example:}
\begin{verbatim}
<exploration>
First, calculate the total cost of the apples: 
5 apples * $2/apple = $10.
Next, subtract the total cost from the amount paid: 
$20 - $10 = $10.
</exploration>
John spends $10 in total. The change is $20 - $10 = $10.
\boxed{10}

<summary>
This solution directly calculated the total cost and 
change. Compared to other paths, it avoided 
unnecessary complications.
</summary>

success
\end{verbatim}

\vspace{0.5em}
\hrule
\vspace{0.5em}
\textbf{Math Problem:} Here is the problem to solve:\\
\texttt{\{question\}}

For solution path \texttt{\{path\_index\}}, please reason step by step, considering insights from the other parallel paths, following above labeling formats and put your final answer within \texttt{\textbackslash boxed\{\}}.
\end{tcolorbox}
\caption{Inference prompt for parallel reasoning. Each thread is aware of its position among parallel paths and is instructed to leverage cross-thread insights while generating its solution, followed by self-assessment and labeling.}
\label{fig:parallel_inference_prompt}
\end{figure}

\clearpage
\section*{NeurIPS Paper Checklist}

\begin{enumerate}

\item {\bf Claims}
    \item[] Question: Do the main claims made in the abstract and introduction accurately reflect the paper's contributions and scope?
    \item[] Answer: \answerYes{}
    \item[] Justification: The abstract and introduction state the paper's scope as a lattice-attention framework for cross-thread reasoning, and the claims are supported by the method, experiments, and ablations in \cref{sec:data,sec:lace,sec:training,sec:main-results}.

\item {\bf Limitations}
    \item[] Question: Does the paper discuss the limitations of the work performed by the authors?
    \item[] Answer: \answerYes{}
    \item[] Justification: Appendix~\cref{appendix:limitations} discusses the main limitation: LACE introduces additional cross-thread communication overhead through lattice attention. The same appendix reports micro-benchmarks in \cref{tab:overhead_17b,tab:overhead_4b}, showing that the routing path is lightweight and the overhead is tolerable in practice.

\item {\bf Theory assumptions and proofs}
    \item[] Question: For each theoretical result, does the paper provide the full set of assumptions and a complete (and correct) proof?
    \item[] Answer: \answerYes{}
    \item[] Justification: Appendix~\cref{sec:analysis} states the conditional-error assumptions used in the analytical comparison between independent sampling and LACE and provides the corresponding derivation. This result is intended as an explanatory analysis rather than a fully general guarantee.

    \item {\bf Experimental result reproducibility}
    \item[] Question: Does the paper fully disclose all the information needed to reproduce the main experimental results of the paper to the extent that it affects the main claims and/or conclusions of the paper (regardless of whether the code and data are provided or not)?
    \item[] Answer: \answerYes{}
    \item[] Justification: The method, baselines, evaluation metrics, data curation procedure, and training setup are described in \cref{sec:data,sec:lace,sec:training,sec:main-results} and the appendix. Appendix~\cref{appendix:data_stats,appendix:implementation} further provides data statistics and implementation details.

\item {\bf Open access to data and code}
    \item[] Question: Does the paper provide open access to the data and code, with sufficient instructions to faithfully reproduce the main experimental results, as described in supplemental material?
    \item[] Answer: \answerNo{}
    \item[] Justification: We do not include a public code and data release with the anonymous submission. The paper nevertheless provides the model architecture, training recipe, data curation procedure, prompts, datasets used, and evaluation protocol needed to support reproducibility, and we plan to release code and curated data after the review process.

\item {\bf Experimental setting/details}
    \item[] Question: Does the paper specify all the training and test details (e.g., data splits, hyperparameters, how they were chosen, type of optimizer) necessary to understand the results?
    \item[] Answer: \answerYes{}
    \item[] Justification: The experimental setup is described in \cref{sec:main-results} and Appendix~\cref{appendix:data_stats,appendix:implementation}, including model sizes, baselines, datasets, thread count, training stages, optimizer, learning rates, checkpoint choices, data statistics, and inference settings.

\item {\bf Experiment statistical significance}
    \item[] Question: Does the paper report error bars suitably and correctly defined or other appropriate information about the statistical significance of the experiments?
    \item[] Answer: \answerNo{}
    \item[] Justification: We do not report formal confidence intervals or error bars for the main tables. We instead report controlled comparisons, ablations, and additional sensitivity/scaling analyses; repeated-run statistical uncertainty remains a limitation, especially for small benchmarks such as AIME.

\item {\bf Experiments compute resources}
    \item[] Question: For each experiment, does the paper provide sufficient information on the computer resources (type of compute workers, memory, time of execution) needed to reproduce the experiments?
    \item[] Answer: \answerNo{}
    \item[] Justification: Appendix~\cref{appendix:overhead} reports decoding micro-benchmark hardware and memory/latency overhead, but the current submission does not fully specify wall-clock time and hardware for every training run. We provide model sizes, training steps, and implementation details, and will include fuller compute accounting in a release version.
    
\item {\bf Code of ethics}
    \item[] Question: Does the research conducted in the paper conform, in every respect, with the NeurIPS Code of Ethics \url{https://neurips.cc/public/EthicsGuidelines}?
    \item[] Answer: \answerYes{}
    \item[] Justification: The work uses public benchmarks and model/data resources for machine learning research, and we are not aware of any deviation from the NeurIPS Code of Ethics. The submission preserves author anonymity and does not involve human subjects or private personal data collection.

\item {\bf Broader impacts}
    \item[] Question: Does the paper discuss both potential positive societal impacts and negative societal impacts of the work performed?
    \item[] Answer: \answerYes{}
    \item[] Justification: Appendix~\cref{appendix:broader_impacts} discusses potential benefits for reliable reasoning and possible dual-use risks from stronger reasoning and agentic capabilities. It also notes that responsible deployment should follow safety evaluation, access-control, and misuse-monitoring practices where appropriate.
    
\item {\bf Safeguards}
    \item[] Question: Does the paper describe safeguards that have been put in place for responsible release of data or models that have a high risk for misuse (e.g., pre-trained language models, image generators, or scraped datasets)?
    \item[] Answer: \answerNA{}
    \item[] Justification: The anonymous submission does not release a high-risk pretrained model, image generator, scraped dataset, or other asset requiring special safeguards. Any future release of checkpoints or curated data will follow appropriate usage terms and safety considerations.

\item {\bf Licenses for existing assets}
    \item[] Question: Are the creators or original owners of assets (e.g., code, data, models), used in the paper, properly credited and are the license and terms of use explicitly mentioned and properly respected?
    \item[] Answer: \answerNo{}
    \item[] Justification: We cite the existing models, datasets, and benchmarks used in the paper, including Qwen3, DAPO, AIME, LiveBench, TALES/TextWorld, and related baselines. The current manuscript does not explicitly list the license terms for every existing asset, and we will add a complete license inventory in the released materials.

\item {\bf New assets}
    \item[] Question: Are new assets introduced in the paper well documented and is the documentation provided alongside the assets?
    \item[] Answer: \answerNA{}
    \item[] Justification: The anonymous submission does not release new datasets, checkpoints, or software artifacts as standalone assets. The paper documents the proposed method and data curation procedure; any future release will include documentation, licenses, and usage notes.

\item {\bf Crowdsourcing and research with human subjects}
    \item[] Question: For crowdsourcing experiments and research with human subjects, does the paper include the full text of instructions given to participants and screenshots, if applicable, as well as details about compensation (if any)? 
    \item[] Answer: \answerNA{}
    \item[] Justification: The work does not involve crowdsourcing experiments or research with human subjects.

\item {\bf Institutional review board (IRB) approvals or equivalent for research with human subjects}
    \item[] Question: Does the paper describe potential risks incurred by study participants, whether such risks were disclosed to the subjects, and whether Institutional Review Board (IRB) approvals (or an equivalent approval/review based on the requirements of your country or institution) were obtained?
    \item[] Answer: \answerNA{}
    \item[] Justification: The work does not involve crowdsourcing experiments or research with human subjects, so IRB approval or equivalent review is not applicable.

\item {\bf Declaration of LLM usage}
    \item[] Question: Does the paper describe the usage of LLMs if it is an important, original, or non-standard component of the core methods in this research? Note that if the LLM is used only for writing, editing, or formatting purposes and does \emph{not} impact the core methodology, scientific rigor, or originality of the research, declaration is not required.
    \item[] Answer: \answerYes{}
    \item[] Justification: LLMs are central to the research: the proposed method modifies Qwen3-based models, and the data curation pipeline uses Qwen3-235B-A22B to generate diverse multi-thread reasoning traces as described in \cref{sec:data} and Appendix~\cref{appendix:implementation}. Any use of LLMs for writing or editing is separate from the scientific methodology.

\end{enumerate}

\end{document}